\documentclass[preprint,11pt,authoryear]{elsarticle}

\usepackage[margin=2.5cm]{geometry}
\usepackage{longtable,multirow}
\usepackage{booktabs}
\usepackage{colortbl}

\usepackage{pgfplots}
\pgfplotsset{compat=1.18} 
\usepackage{filecontents,ulem}
\usepackage{subcaption}
\usepackage{pgfplotstable} 
\usepackage{caption}
\usetikzlibrary{external}
\definecolor{lgray}{gray}{0.85}

\usepackage{amssymb}
\usepackage{amsthm}

\usepackage[ruled,vlined]{algorithm2e} 
\usepackage{xcolor}
\usepackage{euscript}
\usepackage{verbatim}
\usepackage{threeparttable}
\usepackage{setspace}
\usepackage{colortbl} 
\usepackage{amsmath}
\usepackage{natbib}
\usepackage{adjustbox}
\usepackage{multirow}
\usepackage{hyperref}

\usepackage{nth}
\usepackage{graphicx}
\usepackage{blindtext}
\usepackage{subcaption}
\usepackage{svg}

\usepackage{tabularx}
\usepackage{array}

\usepackage{makecell}
\usepackage{threeparttable}
\usepackage{booktabs}
\usepackage{pdflscape}

\AtBeginEnvironment{thebibliography}{\onehalfspacing}

\newtheorem{mydef}{Definition}
\newtheorem{myclaim}{Claim}

\journal{European Journal of Operational Research}

\begin{document}
\onehalfspacing

\definecolor{lightgray}{gray}{0.9}
\definecolor{lightergray}{gray}{0.95}
\begin{frontmatter}



\title{An Efficient Hybridization of Graph Representation Learning and Metaheuristics for the Constrained Incremental Graph Drawing Problem}


\author[inst1,inst2]{Bruna Cristina Braga Charytitsch}

\affiliation[inst1]{organization={Universidade Federal de São Paulo (Unifesp)},
            addressline={Av. Cesare Mansueto Giulio Lattes, 1201 - Eugênio de Mello}, 
            city={São José dos Campos},
            postcode={12247014}, 
            state={SP},
            country={Brazil}}

\author[inst2]{Mari\' a Cristina Vasconcelos Nascimento}

\affiliation[inst2]{organization={Instituto Tecnológico de Aeronáutica (ITA)},
            addressline={Praça Marechal Eduardo Gomes, 50 - Vila das Acacias}, 
            city={São José dos Campos},
            postcode={12228-900}, 
            state={SP},
            country={Brazil}}

\begin{abstract}
Hybridizing machine learning techniques with metaheuristics has attracted significant attention in recent years. Many attempts employ supervised or reinforcement learning to support the decision-making of heuristic methods. However, in some cases, these techniques are deemed too time-consuming and not competitive with hand-crafted heuristics. This paper proposes a hybridization between metaheuristics and a less expensive learning strategy to extract the latent structure of graphs, known as Graph Representation Learning (GRL). 
\textcolor{black}{For such, we approach the Constrained Incremental Graph Drawing Problem (C-IGDP), a hierarchical graph visualization problem. There is limited literature on methods for this problem, for which Greedy Randomized Search Procedures (GRASP) heuristics have shown promising results. In line with this, this paper investigates the gains of incorporating GRL into the construction phase of GRASP, which we refer to as Graph Learning GRASP (GL-GRASP). }
In computational experiments, we first analyze the results achieved considering different node embedding techniques, where deep learning-based strategies stood out. The evaluation considered the primal integral measure that assesses the quality of the solutions according to the required time for such. According to this measure, the best GL-GRASP heuristics demonstrated superior performance than state-of-the-art literature GRASP heuristics for the problem. {\color{black} A scalability test on newly generated denser instances under a fixed time limit further confirmed the robustness of the GL-GRASP heuristics.}

\end{abstract}


\end{frontmatter}

\section{Introduction}
\label{sec:introduction}

Deep learning (DL) has received significant attention in the last decade due to the outstanding results of the computational models based on its paradigm. A relatively new DL method to extract features of graphs, known as Graph Representation Learning (GRL), consists of techniques to extract structural information about a graph.  This information is provided through encoders that map the graph structure to node embeddings, i.e. vectors of features. These strategies act locally to define the graph structure making the computational burden of deep learning less expensive \citep{hamilton2018representation2.4, Hamiltonbook}.
Data predictors that consider the node embeddings significantly outperform state-of-the-art strategies found in the literature in different tasks: classifying protein roles in protein-protein interactions networks, multi-label classification, and link prediction, among others. 

Despite the application of GRL in different settings of graph-related tasks, to our knowledge, Graph Representation Learning to support the decision-making of combinatorial optimization problems (COPs) has not been investigated yet. \cite{bengio2021machine} give an overview of the efforts made by machine learning (ML) and operations research communities to solve COPs. One of the categories highlighted by the authors is employing ML techniques alongside optimization algorithms -- which consists of using ML models for decision-making during the search within optimization algorithms. 
In line with this, hybridizing ML techniques with metaheuristics (MHs) to tackle COPs fits this category when the strategies are cooperative.
 
MHs are approximate optimization methods that solve complex problems adequately within reasonable computational time. Various studies have contributed to this hybridization, introducing new taxonomies to elucidate these interactions \citep{Song2019_1.3, Talbi2021}. \cite{Karimi} provide a comprehensive review of recent progress in approaches that combine ML with MHs for COPs. They delve into aspects such as algorithm selection, fitness evaluation, initialization, evolution, parameter setting, and cooperation, citing a diverse range of COPs involved in these endeavors.

This paper introduces a hybridization of DL with MHs to approach a graph optimization problem known as Graph Drawing (GD).
\textcolor{black}{The graph drawing is a geometric representation problem where GRL may perform better.}

In particular, the Constrained Incremental Graph Drawing problem (C-IGDP) proposed by \cite{napoletano2019} considers layered networks and is the problem for which hybrid methods were proposed. In this problem, a $\Lambda$-partite graph has an initial 2D configuration, where the vertices (or nodes) of the $\Lambda$ layers are displayed in $\Lambda$ columns. Incremental vertices must be positioned so that the number of arc crossings is minimized. A few heuristics were proposed to solve the C-IGDP, like the Greedy Randomized Search Procedure (GRASP) \textcolor{black}{\citep{grasp}} and the Tabu Search \textcolor{black}{\citep{Glover_Tabu}}. In the review conducted by \cite{Karimi} on the joint use of ML and MHs in solving COPs, no GD problem is in the comprehensive list of COPs approached through these methods.

The methods introduced in this paper use information from GRL to guide the search for solutions. They are based on the literature heuristics for the C-IGDP but adapted so they use the graph embedding information. Computational experiments considered four distinct methods for generating node embeddings: two based on DL, one spectral method, and one based on matrix factorization. Stochastic node embedding strategies had two different learning GRASP versions. The approaches proved promising, particularly those involving DL, that outperformed state-of-the-art methods in denser \textcolor{black}{graphs}.

The primary contributions of this paper are summarized next.

\begin{itemize}
    \item \textbf{Proposition of a new form to combine ML with MHs to solve the C-IGDP:} The chosen problem was a visualization task in graph problems, for which we identified potential in using the embeddings obtained by GRL to help the construction of solutions.
    \item \textbf{Advances on the state-of-the-art C-IGDP heuristic methods:} This study contributes to the advancement of the C-IGDP,  a recently proposed problem that lacks further exploration. We achieved competitive results through the learning-based GRASP, investigating the potential of GRL for COPs and filling a knowledge gap.
    \item \textbf{Comparison, implementation, and information:} Detailed description of the results with optimal solutions are provided, along with the implementation and a comparative analysis of six introduced variations of the Graph Learning GRASP (GL-GRASP) heuristics with the literature methods. This analysis offers a comprehensive view of different approaches and a valuable reference for evaluating the effectiveness of the proposals. All implementations and optimal solutions are available in this paper.
    \item \textbf{Integration between ML and optimization algorithms:} This work is at the intersection of ML and optimization algorithms, combining DL with MHs and exploring the diversity of embedding generation methods to provide an efficient solution for the C-IGDP and advance knowledge in this multidisciplinary area.
\end{itemize}

\textcolor{black}{The remainder of this paper is organized as follows. Section~\ref{sec:sample1} reviews the literature on incremental graph drawing, the application of machine learning and metaheuristics to combinatorial optimization problems, and the use of graph representation learning for heuristic guidance. Section~\ref{sec:sample2} formally defines the Constrained Incremental Graph Drawing Problem, presents its mathematical model, and discusses the baseline GRASP heuristics. Section~\ref{sec:sample3} details the proposed GL-GRASP heuristics, explaining the node embedding techniques used and the GRL-based construction phase. Section~\ref{sec:sample4} outlines the setup for the computational experiments, including both benchmark and newly generated data, parameter settings, and assessment metrics. It also provides a thorough analysis of the results obtained from evaluating the GL-GRASP heuristics, along with a comparative analysis of methods from the literature. Finally, Section~\ref{sec:sample5} summarizes the key conclusions of this work and suggests potential directions for future research.}


\section{\textcolor{black}{Related Works}}
\label{sec:sample1}

{\color{black}This section presents a literature review of studies related to this investigation. Section~\ref{sec:sample1_a} discusses the literature on the {\color{black}Incremental Graph Drawing}. Section~\ref{sec:sample1b} provides a review of the synergy between machine learning techniques and metaheuristics for solving COPs. Finally, Section~\ref{sec:sample2_GRL} presents how GRL has been used to guide decisions within optimization algorithms.}

\subsection{\textcolor{black}{ Incremental Graph Drawing}}
\label{sec:sample1_a}
\textcolor{black}{ Graph Drawing (GD) is a well-established research subject focused on the automatic layout of graphs \citep{kaufmann3.0}. Although earlier contributions exist, the field was formally introduced by \cite{tamassia1995graph}. To encourage further exploration, \cite{Binucci:3.1} presented compelling motivations for investigating GD, including foundational research and applications in information visualization, software engineering, model-based design, automated cartography, social sciences, and molecular biology. }

\textcolor{black}{ A central aspect of GD is considering aesthetic criteria to improve the readability and utility of graph layouts while balancing construction costs. Key criteria include minimizing edge crossings and bends, maximizing symmetry and angular resolution, reducing area usage and edge length, and achieving uniform vertex distribution \citep{batista3.3}. Among them, minimizing edge crossings is particularly important, as it enhances graph readability \citep{ware3.4}. This issue is formalized in the Edge Crossing Minimization Problem (ECMP) \citep{Junger1997}, an NP-hard combinatorial problem that is central to many approaches in graph drawing.}

\textcolor{black}{
Formulated as a linear ordering problem, the ECMP seeks an optimal vertex arrangement across layers to minimize arc crossings. Despite decades of research, it remains computationally demanding, especially as the number of layers and vertices grows. The typical approach involves a hierarchical layout of directed acyclic graphs, where vertices are placed on equidistant layers, and edges or arcs connect vertices between layers. There are two main strategies for the ECMP: the layer-by-layer sweep and the multi-layer sweep. The former focuses on minimizing crossings between adjacent layers of a bipartite graph, employing approaches such as one-sided, two-sided, and centered 3-level crossing minimization \citep{Gunther, Junger1997, Bachmaier2010}. In contrast, the multi-layer sweep seeks a global solution by reordering vertices across all layers, often by iteratively applying bipartite minimization techniques to successive layers.}

\textcolor{black}{Solutions to the ECMP include both heuristics and exact algorithms. Among them, the method proposed by \citep{Sugiyama3.9} is one of the most widely used, consisting of four steps: cycle removal, layer assignment, crossing minimization, and coordinate assignment.  
The barycenter heuristic \citep{Makinen}, which positions vertices near their neighbors to reduce crossings, is particularly popular due to its speed, simplicity, and effectiveness \citep{kaufmann3.0}. Other heuristics explored in the literature include the median, greedy switch, split, greedy insertion, stochastic, assignment, and sifting methods. 
\cite{Junger1997} also present an exact algorithm for two-layer crossing minimization, which is computationally feasible when one layer is fixed or the instance is small (e.g., up to 15 nodes in the smaller layer).}

\textcolor{black}{While these methods target static graph layouts, dynamic scenarios introduce new challenges, particularly when modifying an existing drawing with additional (incremental) vertices and edges. In such cases, preserving the original structure of the layout, known as maintaining the user’s mental map, is crucial for interpretability. As emphasized by \cite{eades1991preserving}, minimizing changes to the original layout helps users in tracking graph evolution. This need leads to the Incremental Graph Drawing Problem (IGDP), which focuses on minimizing new arc crossings while preserving the relative order of non-incremental vertices as updates occur\footnote{This means that the precedence relation between non-incremental vertices from the same layer must remain unchanged.}}.
\cite{MARTI20011287} proposed a branch-and-bound method and a GRASP heuristic for the IGDP in the context of 2-layered graphs, called Incremental Bipartite Drawing Problem (IBDP). As in the multi-layered case, the goal of the IBDP is to preserve the relative position of the vertices regarding the initial drawing while minimizing the number of arc crossings. The vertex degree-based heuristic effectively dealt with medium-sized and large sparse instances. The exact solution solved instances with up to 32 vertices.

Later, \cite{MARTI20181} renamed the IBDP to Dynamic Bipartite Drawing Problem (DBDP), following contemporary literature on similar graph drawing problems. The authors introduced a hybrid algorithm that combines an adaptation of the constructive phase of the GRASP proposed by \cite{MARTI20011287} with a short-term memory Tabu Search. Additionally, they hybridized the method with a path-relinking strategy, achieving even better solutions to the problem.

\cite{Sanchez2017} extended the GRASP framework proposed by \cite{MARTI20011287} to multiple layered case. They also introduced a new methodology for the multi-layered incremental graph drawing problem by hybridizing the Variable Neighborhood Search (VNS) heuristic with the \textcolor{black}{Scatter Search} methodology (SS), \textcolor{black}{referred to as VNSS}. The proposed method has shown promise by preserving key characteristics of the VNS and SS while also demonstrating an excellent balance between intensification and diversification. Out of the 48 tested instance configurations, the method yielded the best solutions in 41 cases, particularly excelling in graphs with more incremental vertices.

\cite{PENG2020183} proposed an iterated solution-based Tabu Search algorithm with adaptive memory for the incremental and decremental variants of the DBDP. The study incorporates a restricted neighborhood structure and an efficient hash function to promote search intensification. Furthermore, diversification was tackled through an adaptive perturbation mechanism. The experiments with over 1000 instances demonstrated high competitiveness compared to existing alternatives. Moreover, the research highlights that solution-based memory structures using hash functions can significantly improve the speed and efficiency compared to traditional attributional structures, particularly in search intensification. 

More recently, \cite{PENG2024121477}  introduced a heuristic method that combines the Tabu Search with multiple neighborhood structures (MNSB-TS) to address the multi-layered IGDP. The suggested method offers a good balance between time and space besides achieving solid performance compared to the best-performing heuristics proposed for the problem (GRASP, VNSS, and SS) and the Gurobi solver. It presents an effective search procedure with a simple implementation by incorporating a multiple neighborhood mechanism, leading to increased diversification in the IGDP resolution.

\textcolor{black}{\cite{napoletano2019} proposed a more constrained variant of the IGDP, the so-called Constrained Incremental Graph Drawing Problem (C-IGDP). In addition to preserving the relative precedence order of the original vertices from the same layer, the absolute position of each original vertex is constrained to remain within a fixed distance from its position in the original layout. The objective of the C-IGDP is to minimize the number of edge crossings. Despite being a relatively recent proposal, the C-IGDP remains underexplored in the literature. The authors presented GRASP heuristics with distinct constructive phases, and a Tabu Search heuristic. The results showed that all heuristics were competitive across multiple instance sets. Moreover, they observed that diversification in search-based methods was achieved through memory structures and semi-random designs, which varied according to the specific characteristics of the problem. In their experiments, both GRASP and memory-based approaches exhibited similar performance. The C-IGDP and the GRASP heuristics are discussed in detail in Section~\ref{sec:sample2}.}


\textcolor{black}{In summary, methods for solving the ECMP and its dynamic variants have primarily relied on heuristics and metaheuristics, which have proven effective in handling graphs of varying sizes and complexities. Recent contributions to the IGDP and C-IGDP reflect this trend by tackling dynamic aspects while preserving the integrity of graph representations. These approaches serve as a foundation for the strategies explored in this paper. Motivated by recent advances, this study proposes integrating ML into MHs to tackle the C-IGDP more effectively. The following section provides an overview of how ML and MH have been employed in solving COPs.}

\subsection{Machine Learning and Metaheuristics for COPs}
\label{sec:sample1b}

The integration of ML with MHs has gained significant traction in solving Combinatorial Optimization Problems \citep{KALLESTAD2023446, Karimi, Talbi2021, bengio2021machine, Song2019_1.3, RePEc_Talbi}.
MHs are well-known for their ability to explore large solution spaces and deliver high-quality solutions within a reasonable time. ML complements these methods by enhancing convergence speed, robustness, and overall solution quality.

In their review, \cite{Song2019_1.3} classify ML-optimization interactions into two categories: self-interactions, where learning or optimization improves itself, and dual interactions, where both mutually benefit. Expanding on this, \cite{bengio2021machine} define three main paradigms: end-to-end learning, where models directly output solutions from raw input; learning to configure, which automates parameter tuning; and ML alongside optimization, where ML assists the optimization process dynamically, e.g., by guiding neighborhood selection, diversification, or solution quality estimation.
From a structural perspective, \cite{Talbi2021} categorizes the integration of ML into MHs across three levels: problem-level, high-level, and low-level. These correspond respectively to modeling and approximating the optimization problem, guiding the selection or coordination of MHs, and tuning algorithmic parameters. 

Recent advances in integrating ML into MHs for solving COPs have focused on enhancing existing algorithms rather than designing entirely new ones \citep{Peres}. ML techniques have been successfully applied to various components of MHs, such as algorithm selection, fitness evaluation, initialization, and parameter tuning, resulting in more robust and efficient search processes \citep{Karimi}. However, while a wide range of COPs has been explored, Graph Drawing problems and their variants remain largely unaddressed in this hybrid context. Given the high computational complexity of GD problems, integrating ML into MHs represents a promising direction, particularly for dynamic or constrained scenarios.

Despite ongoing progress, most studies continue to rely on traditional ML techniques like Support Vector Machines (SVM), k-means, and k-Nearest Neighbors (k-NN) \citep{Song2019_1.3, Karimi}. While useful in specific contexts, these methods often struggle with scalability and generalization in more complex environments. Deep Learning frameworks have the potential to address these limitations due to their capacity to model intricate patterns. Real-world applications in logistics, transportation, and network design underscore the growing need for adaptive, learning-driven optimization. Within this landscape, dynamic and incremental GD problems are particularly compelling yet remain underexplored in the ML-MH hybridization efforts.

In GD, ML has primarily been applied through end-to-end learning approaches, such as Graph-LSTM \citep{wang3.43} and DeepGD \citep{Wang2021_}), and learning-based configuration strategies. Recent advances leverage deep learning, particularly Graph Neural Networks (GNNs). For example, \citep{Tiezzi2022} uses a GNN to create graph layouts by learning aesthetic optimization criteria from established methods. Similarly, \citep{Giovannangeli2024} introduced a GNN-based framework for GD to enhance adaptability through transfer learning, using GNNs and convolutions without predefined ground truth. These approaches improve layout quality and address the limitations of classical methods. However, their integration into the core search process— within the ML Alongside Optimization paradigm — remains limited.

This work seeks to bridge that gap by combining ML with classical metaheuristics methods to solve the C-IGDP, using learned representations to generate or evaluate solutions and guide the search process actively. To this end, the following section introduces Graph Representation Learning, a framework for encoding graph structures into vector embeddings, to steer the metaheuristic optimization process.

\subsection{\textcolor{black}{Graph Representation Learning as Heuristic Guidance}}
\label{sec:sample2_GRL}

{\color{black}Graph Representation Learning (GRL) is a research area that explores the use of machine learning to solve problems modeled by graphs. A core technique in GRL is graph embedding, which maps nodes, subgraphs, or entire graphs into low-dimensional vector spaces while preserving structural properties. Graph embeddings enable the application of standard ML techniques and are used in tasks such as node classification, link prediction, clustering, and graph visualization \citep{hamilton2018representation2.4}. 

 While the use of graph embeddings in such downstream tasks is well established, particularly in supervised and self-supervised settings \citep{Fenxiao2020, Hoang, Wei_Ju_2024,shima2024}, their application as heuristic guidance in optimization algorithms remains underexplored. The survey by \cite{Hoang}, for example, provides a comprehensive taxonomy of GRL approaches despite not covering studies that integrate pre-computed embeddings into classical metaheuristics. Most existing literature that combines GRL with optimization regards end-to-end learning pipelines where embeddings are dynamically updated during the search. 
 
 \cite{CHEN2022} presented one of the few studies to explore this integration, an Ant Colony Optimization (ACO) and graph embedding hybrid (ACO-GE), to address the problem of suppressing the spread of negative influence in social networks under cost constraints. In their approach, node embeddings are generated in a preprocessing stage using the DeepWalk \citep{Perozzi2014} algorithm, which captures structural roles through truncated random walks. These embeddings are then used to compute pairwise Euclidean distances between nodes, which are normalized and incorporated as heuristic factors during the ACO search process. The embeddings remain fixed throughout the optimization, serving as static guidance to enhance the metaheuristic's decision-making by embedding structural information directly into its path construction mechanism.

Another contribution comes from \cite{LIU2022}, who tackle the multi-objective multigraph shortest path problem, which is a complex extension of classical pathfinding where multiple parallel edges may exist between nodes, each associated with non-dominated cost vectors across several objectives. To address the computational challenges inherent in such problems, the authors introduce an extension of the A* search tailored for multi-objective multigraphs. The Node2Vec algorithm generates the node embeddings  \citep{Grover} to predict heuristic cost estimates for each objective via pre-trained neural networks. During the execution of the multi-objective A* search, the learned heuristic cost estimates guide the search process, enhancing the identification of Pareto-optimal paths. As in ACO-GE, the node embeddings are pre-computed and remain fixed throughout the search, functioning as static structural encodings that guide and enhance the performance of the underlying heuristic algorithm.

 \cite{li2018} use graph convolutional networks to predict node probabilities for optimal solutions, guiding a tree search. Unlike precomputed embeddings, embeddings are recalculated during the search, contrasting with the fixed-embedding strategies in this work. 
 
 While several studies at the intersection of GRL and combinatorial optimization leverage learned embeddings to improve solution quality, most adopt end-to-end frameworks in which embeddings are generated and refined during the optimization process. For instance, \cite{dai2018} and \cite{gasse2019} employ neural networks that dynamically generate node or variable embeddings at each decision point, tightly integrating the representation learning with the search procedure.
Similarly, \cite{ye2023} replace handcrafted heuristics in ACO with neural policies trained via reinforcement learning, where embeddings are updated jointly with the metaheuristic behavior. In these cases, embeddings are not precomputed nor used as static inputs to external optimization routines but act as internal, trainable components of the decision-making process. Consequently, they fall outside the scope of this work, which focuses on settings where graph embeddings are computed independently and subsequently used to inform or enhance traditional metaheuristics as fixed heuristic signals.

The following section presents the C-IGDP and its formulation.
}


\section{Problem and Formulation}
\label{sec:sample2}

\begin{figure}[!h]
\centering
\includegraphics[scale=1]{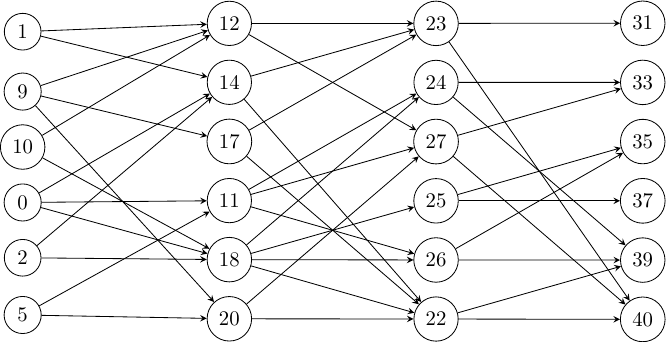}
\caption{Visual depiction of a hierarchical graph with four layers, each of them consisting of six nodes.} \label{fig:example-graph1}
\end{figure}

Consider a drawing $\mathop{\mathcal{D} = (G, \Pi_{0})}$ as the hierarchical representation of a multi-layer graph. In this paper, $\mathop{G=(V, A, \Lambda)}$ denotes the graph and its elements, where $V^{\lambda}$ represents the set of vertices of layer $\lambda$ ($\mathop{V= V^{1} \cup ... \cup V^{\Lambda}}$), $A^{\lambda}$ denotes the set of arcs leaving vertices from layer $\lambda$ ($\mathop{A = A^{1} \cup ... \cup A^{\Lambda-1}}$) and $\Lambda$ indicates the total number of layers.  
Figure~\ref{fig:example-graph1} depicts the drawing $\mathcal{D}$ of a 4-layer graph $G$, where the 24 vertices are distributed across layers and arcs link vertices in successive layers. 

Minimizing the number of arc crossings in a drawing is one of the necessary requisites to define a good layout. To identify the number of arc crossings, consider $\mathop{\Pi_{0} = \{ \pi^{1}_{0}, \pi^{2}_{0},\ldots \pi^{\Lambda}_{0} \}}$, where $\pi^{\lambda}_{0}: V^\lambda \mapsto \{1, 2, \ldots, |V^\lambda|\}$ is a function that indicates the position of the vertices of a given layer $\lambda$. In Figure~\ref{fig:example-graph1}, $\pi^{1}_{0}(1)=\pi^{2}_{0}(12)=\pi^{3}_{0}(23)=\pi^{4}_{0}(31)=1$, indicating that vertices 1, 12, 23 and 31 are in the first position of the corresponding layer. On the other hand, $\pi^{1}_{0}(9)=\pi^{2}_{0}(14)=\pi^{3}_{0}(24)=\pi^{4}_{0}(33)=2$ which means that vertices 9, 14, 24 and 33 are in the second position of their layer. We say that arcs ($a,b$) and ($c,d$), where $a, c\in V^{\lambda-1}$ and $b,d \in V^{\lambda}$ connecting consecutive layers cross if either $\pi^{{\lambda-1}}_{0}(a)<\pi^{{\lambda-1}}_{0}(c)$ and $\pi^{\lambda}_{0}(b)>\pi^{\lambda}_{0}(d)$ or $\pi^{\lambda-1}_{0}(a)>\pi^{\lambda-1}_{0}(c)$ and $\pi^{\lambda}_{0}(b)<\pi^{\lambda}_{0}(d)$.  In the example, arcs (18, 26) and (20, 27) cross because $\pi^{2}_{0}(18) = 5 < \pi^{2}_{0}(20) = 6$ and $\pi^{3}_{0}(26) = 5 >\pi^{3}_{0}(27) = 3$.

Now, given an initial hierarchical configuration of the vertices and arcs, like in the example presented in Figure~\ref{fig:example-graph1}, consider that new vertices and arcs need to be added to the same graph. According to \cite{eades1991preserving}, keeping the original structure as much as possible is more intuitive for an analyst to infer about the visualization of this network applied to a specific context. Graphs are models for various real-world problems, including affiliation networks, project scheduling, and software visualization. Nowadays, specialists can access systems designed for analyzing and dynamically interacting with these problems. In dynamic diagrams, for instance, where there are connections between elements, specialists must retain a clear understanding of the overall layout (the user's mental map) even after minor modifications, such as adding or removing links or elements. It is not ideal for a diagram to be completely modified due to minor alterations.

\begin{figure}[!h]
\centering
\includegraphics[scale=1]{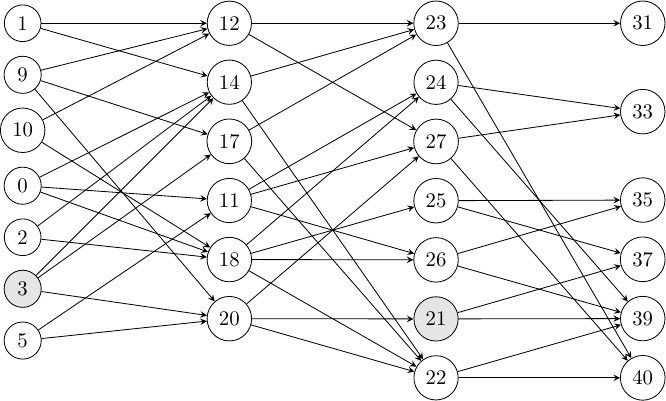}
\caption{Visualization of a hierarchical graph: transitioning from four layers with six vertices each to an expanded structure with added vertices and arcs.} \label{ffig:example-graph2}
\end{figure}

In line with this, the goal of the Incremental Graph Drawing Problem (IGDP) is to reduce the number of arc crossings by strategically rearranging vertices when introducing new elements to the graph. Simultaneously, efforts are made to maintain the relative 
positions of the original vertices, which will be further discussed later. When incorporating a set of incremental vertices $\widehat{V}=\widehat{V}^1\cup \widehat{V}^2 \cup \ldots \widehat{V}^{\Lambda}$ and their corresponding arcs $\widehat{A}=\{\widehat{A}^1\cup \widehat{A}^2 \ldots \widehat{A}^{\Lambda-1}\}$ into a hierarchical graph $G$, an incremental graph denoted as $\mathop{IG=(IV, IA, \Lambda)}$ must be created. In this paper, $IV^\lambda = V^\lambda \cup \widehat{V}^\lambda$, $\forall \lambda \in\{1, 2, \ldots, \Lambda\}$, and $\mathop{IV= \{IV^1\cup IV^2 \ldots\cup IV^\Lambda\}}$ and $IA^\lambda = A^\lambda \cup\widehat{A}^\lambda$,  $\forall \lambda \in\{1, 2, \ldots, \Lambda-1\}$, and $IA= \{IA^1 \cup IA^2 \ldots \cup IA^{\Lambda -1}\}$. It is noteworthy that the total number of layers remains unchanged. In addition, $\mathop{\Pi = \{ \pi^{1}, \pi^{2},\ldots, \pi^{\Lambda} \}}$ represents the new set of position functions. Figure~\ref{ffig:example-graph2} shows a final layout after adding the nodes $\{3, 21\}$ and arcs $\{(3,14), (3,17), (3,20), (20,21), (21,37), (21,39)\}$. In this layout, the original nodes and arcs have not changed their relative positions. The number of arc crossings increased from 65 to 86 after the addition of new elements.

The IGDP was proposed in \cite{Sanchez2017} founded in the so-called Crossing Minimization Problem \citep{Junger1997}, a classic problem in graph theory whose goal is finding an arrangement of arcs that minimizes the number of their crossings. The Constrained Incremental Graph Drawing Problem (C-IGDP) arises in this context as a restricted version of IGDP where efforts are made to maintain the relative and absolute positions of original nodes \citep{napoletano2019} and is better explained next.

\subsection{Mathematical Model}

The goal of the Constrained Incremental Graph Drawing Problem is to find a drawing represented by $\mathcal{I} \mathop{=(IG, \Pi)}$ that not only minimizes the number of arc crossings but also preserves the original relative and absolute ordering of vertices $\mathop{\nu \in V}$ in $\mathcal{D}$. More formally, to understand the concept of relative and absolute ordering, let us define the precedence relationship between a pair of nodes in the same layer. 

\begin{mydef} (Relative precedence)
 Let $\mathcal{D}=(G,\Pi_0)$ be a drawing. A vertex \(i \in L_{\lambda}\) precedes vertex \(j\in L_{\lambda}\), denoted by \(i \prec j\), if $\pi^{\lambda}_0(i) < \pi^{\lambda}_0(j)$. \label{relativeprecedence}    
\end{mydef}

\begin{myclaim}Let $\mathcal{I} \mathop{=(IG, \Pi)}$ be an incremental drawing of $\mathcal{D}=(G,\Pi_0)$. For $\mathcal{I}$ to meet the relative ordering constraint, $\pi^{\lambda}(i) < \pi^{\lambda}(j)$ if and only if $\pi^{\lambda}_0(i) < \pi^{\lambda}_0(j)$.
\end{myclaim}

Besides the relative ordering, one must consider the absolute ordering for $\mathcal{I}$ to be feasible as stated by Claim~\ref{absoluteprecedende}.

\begin{myclaim}Let $\mathcal{I} \mathop{=(IG, \Pi)}$ be an incremental drawing of $\mathcal{D}=(G,\Pi_0)$ and $d$ a natural number referred to as maximum dislocation extent. For $\mathcal{I}$ to meet the absolute positioning constraint, $|\pi_0^\lambda(i)-\pi^\lambda(i)|\leq d, \forall i\in V$.\label{absoluteprecedende}
\end{myclaim}

The C-IGDP aims at finding a minimum arc crossing incremental drawing. The adapted mathematical formulation introduced by \cite{napoletano2019} is shown next. Before presenting the model, consider the following indexes, parameters and decision variables.

\begin{center}
\begin{longtable}{ccp{10cm}}
  Parameters & & \\
  $i,j,l, w,z$ & : & indexes representing nodes;  \\
  $n$&: & number of vertices of the incremental graph;\\
  $\lambda$ & : & index representing the layers;  \\
   $V^{\lambda}$ & : & set of vertices of $G$ in layer $\lambda$; \\ 
   $IV^{\lambda}$ & : & set of vertices of $IG$ in layer $\lambda$; \\ 
   $IA^{\lambda}$ & : & set of arcs of $IG$ with tail in layer $\lambda$ and head in layer $\lambda+1$; \\
   $\Lambda$ & : & number of layers; \\
   $d$ & : & maximum dislocation extent number of positions;  \\
   $\Pi_0$& : & set of position functions of the original graph;\\
  Decision variables &  & \\
  $c_{iwjz}^{\lambda}$ & : & binary variable that receives value 1 if arcs $(i,w)$ and $(j,z)$ cross, and 0, otherwise\\
  $x_{ij}^{\lambda}$ & : & binary variable that receives value 1 if $i\prec j$, and 0 otherwise;  \\
  $\pi^{\lambda}(i)$ & : & integer variable that receives the position of node $i$ in layer $\lambda$. \\ 
 \end{longtable}
\end{center}

The C-IGDP is modeled as follows:

\begin{align}
    \min  \mathop{\sum_{\lambda=1}^{\Lambda-1}} \mathop{\sum_{(i,w),(j, z) \in {IA}^{\lambda}, i \prec j, z \neq w} c_{iwjz}^{\lambda}} & \label{obj1}
\end{align}

subject to:

    {\allowdisplaybreaks \begin{align}
    & \mathop{- c_{iwjz}^{\lambda} \leq x_{w z}^{\lambda+1} - x_{i j}^{\lambda}  \leq c_{iwjz}^{\lambda}} & \forall & \; (i,w), (j, z) \in {IA}^{\lambda}, i \prec j, w \prec z& \label{r2}\\
    &\mathop{1 - c_{iwjz}^{\lambda} \leq x_{z w}^{\lambda+1} + x_{ij}^{\lambda}  \leq 1+ c_{iwjz}^{\lambda}}   & \forall & \; (i,w), (j,z) \in {IA}^{\lambda}, i \prec j, z \prec w& \label{r3}\\
    &\mathop{0 \leq x_{i j}^{\lambda} + x_{j l}^{\lambda}- x_{i l}^{\lambda} \leq 1} & \forall & \; i, j, l \in {IV}^{\lambda}, i \prec j \prec l  &\label{r4} \\
    &\mathop{x_{ij}^{\lambda} + x_{ji}^{\lambda} = 1} & \forall & \; i, j \in {IV}^{\lambda},\mathop{i \prec j} & \label{r5} \\
    &\mathop{x_{ij}^{\lambda} = 1}  & \forall & \; \mathop{i, j \in V^{\lambda}, \pi^{\lambda}_{0}(i) < \pi^{\lambda}_{0}(j)} &\label{r6} \\
    &\mathop{\max\{1, \pi^{\lambda}_{0}(i) - d \} \leq \pi^{\lambda}(i)}  & \forall & \; \mathop{i \in V^{\lambda}}& \label{r7} \\
	&\mathop{\min\{\pi^{\lambda}_{0}(i) + d, |IV^{\lambda}|\} \geq \pi^{\lambda}(i)}   & \forall & \; \mathop{i \in V^{\lambda}} &\label{r8} \\
	&\mathop{\pi^{\lambda}(i) = |IV^{\lambda}| - \sum_{j=1, j \neq i}^{|IV^{\lambda}|} x_{ij}^{\lambda} }  & \forall & \; \mathop{i \in {IV}^{\lambda}} &\label{r9} \\
	&\mathop{x_{i j}^{\lambda}, c_{iwjz}^{\lambda} \in \{ 0,1\}}  & \forall & \; \mathop{(i,w), (j, z) \in {IA}^{\lambda}, i \prec j, z \neq w }  &\label{r10} 
    \end{align}}
    
The objective function~\eqref{obj1} aims to minimize the number of arc crossings. Constraints \eqref{r2} and \eqref{r3} ensure that  variable $\mathop{c_{iwjz}^{\lambda}}$ is 1 when arcs $\mathop{(i,w)}$ and $\mathop{(j,z)}$ cross. The occurrence of a crossing is related to the positions of the vertices in the layers, a relationship explicitly captured by variables $\mathop{x_{wz}^{\lambda+1}}$, $\mathop{x_{zw}^{\lambda+1}}$, and $\mathop{x_{ij}^{\lambda}}$. These variables represent and reflect the dependency on the spatial arrangement of vertices. Constraints \eqref{r4} ensure that the precedence order of three vertices (transitivity) is correctly assigned to variables $x$, meaning that if one vertex precedes another, and the latter precedes a third, the first vertex must precede the third, thus respecting the order relationship of vertices in each layer. Constraints \eqref{r5} enforce that in a given layer $\lambda$,  binary variables $\mathop{x_{ij}^{\lambda}}$ and $\mathop{x_{ji}^{\lambda}}$ are mutually exclusive, i.e. if the former is 1, the latter is 0, and vice-versa. The relative positions of the original vertices are maintained by constraints \eqref{r6}, while constraints \eqref{r7} and \eqref{r8} ensure the preservation of their absolute positions, all within the bounds previously defined by $d$. Constraints \eqref{r9}, which we added in the model, guarantee that $\mathop{\pi^{\lambda}(i)}$ stores the final position of each vertex \(i\) within layer $\lambda$. This equation establishes a connection between  variables $\mathop{\pi^{\lambda}(i)}$ and $\mathop{x^{\lambda}_{ij}}$. It is important to note that the precedence variable returns a unitary value when one vertex precedes another. In a layer with $ |IV^{\lambda}|$ vertices, the final position of a vertex is determined by subtracting from $ |IV^{\lambda}|$ the total number of vertices that the given vertex precedes. The vertex that precedes the greatest number of vertices occupies the first position, while those that precede none are positioned last. Finally, constraints \eqref{r10} define the domain of the variables as binary.

The C-IGDP is closely related to the  Incremental Graph Drawing Problem (IGDP)  \citep{Sanchez2017}. The IGDP consists of the mathematical formulation \eqref{obj1} to \eqref{r6} and \eqref{r10}. 

\textcolor{black}{To heuristically solve the C-IGDP, \cite{napoletano2019} proposed metaheuristics, described in the next section.}

\subsection{\textcolor{black}{GRASP heuristics for the C-IGDP}}\label{sec:const}
{\color{black}GRASP \citep{grasp} is a widely employed iterative metaheuristic {\color{black} \citep{NASCIMENTO2010747,SOUZAALMEIDA2022105804,LAGUNA2025}}, where each iteration has two phases: the semi-greedy construction phase and the local search phase. The construction phase maintains a restricted candidate list (RCL) with the $\mathop{\varphi}\times 100\%$ ($\mathop{\varphi} \in $ [0,1]) best candidates to join the partial solution. The list is updated iteratively after one of the candidates is selected to compose the solution. The process stops when the method finds a feasible solution to be improved by the local search. The process repeats for several iterations, keeping the best overall solution.

\cite{napoletano2019} proposed three GRASP heuristics to solve the C-IGDP, whose key differences lie in the construction phases, namely \( C_1 \), \( C_2 \), and \( C_3 \). }
 
\begin{itemize}
    \item $C_1$: This method uses a strategy by, at first, choosing a high-degree vertex and randomly positioning it in the appropriate layer. An RCL is iteratively updated with the best candidates that are the vertices from the set of vertices to be located whose degree is greater than or equal to a specified percentage ($\mathop{\varphi} \in $ [0,1]) of the highest vertex degree of remaining vertices. The selected vertex is positioned according to the barycenter method \citep{batista3.3}. This barycenter method calculates each vertex's position based on the average position of its neighboring vertices. The goal is to arrange the vertices within a layer in viable positions the closest possible to the calculated average positions.

    \item $C_2$: This method assumes that a partial solution, represented by the original drawing, is given. Then, it works as $C_1$ to place the incremental vertices $\mathop{\widehat{V}}$ through the semi-greedy strategy. 
    \item $\mathop{C_3}$: This method also assumes that a partial solution, represented by the original drawing, is given. A function $\mathop{\varrho(\nu,p)}$ that quantifies the number of additional arc crossings when inserting a vertex $\mathop{\nu}$ at the position $p$ within its layer is considered in the semi-greedy process. In this method, let $\mathop{\varrho(\nu)}$ be the minimum  $\mathop{\varrho(\nu,p)}$ values considering all possible $p$ positions. Therefore, if $\mathop{\nu}$ is selected to join the partial solution, it will be placed in the position that minimizes additional arc crossings. The construction process begins with the entire candidate list $L$ being the incremental vertices, i.e. $\mathop{\widehat{V}}$. The RCL is composed of all vertices to be positioned in position  $\arg{\min_p}\mathop{\varrho(\nu,p)}$  with $\varrho(\nu) $ below or equal to a threshold $\mathop{\xi}$, where $$\mathop{\xi = \min_{\nu \in L} \varrho(\nu) + \varphi \left( \max_{\nu \in L} \varrho(\nu) - \min_{\nu \in L} \varrho(\nu) \right)}$$ A random vertex $\mathop{\nu^*}$ is chosen from RCL. If the position assigned to $\mathop{\nu^*}$ is occupied, the preceding vertices must be shifted upwards. If this movement violates any constraints, the vertex must be inserted into the closest available position possible to the calculated one.

\end{itemize}

{\color{black}The local search procedure is applied to each solution produced in the construction phase. It refines the graph drawing by performing either swap operations, which exchange the positions of two incremental nodes, or insertion operations, which reposition an incremental node within the layout. In the swap movement, two incremental vertices within the same layer exchange positions. Each incremental vertex is systematically evaluated, considering potential swaps with others in the same layer. The method selects the most effective feasible swap to reduce arc crossings and then proceeds to the next incremental vertex. This iterative process continues until no further improvements can be achieved, and the insert movements start.

In the insert moment, each incremental vertex is examined for potential insertions into its preceding positions within a designated layer, ensuring that it complies with constraints related to the positions of the original vertices. If the movement improves the solution, it is implemented. Otherwise, the algorithm iterates through subsequent insertion options to identify the most advantageous one. This iterative process continues to the next vertex in the same layer after each examination. Computational experiments indicated that the GRASP heuristics incorporating $C_2$ and $C_3$ as construction phases achieved the best performance.}


\section{\textcolor{black}{ GL-GRASP heuristics}}
\label{sec:sample3}

 {\color{black}This section presents the proposed Graph Learning GRASP (GL-GRASP) heuristics, which consist of the GRASP heuristics introduced by \cite{napoletano2019} adapted to incorporate learning in their construction phases. Before presenting the proposed metaheuristics, we provide an overview of the GRL algorithms used to extract node embeddings from the input data. Then, the construction phase of the GL-GRASP heuristics is thoroughly described. \textcolor{black}{The local search phase in the proposed GL-GRASP heuristics is identical to that described by \cite{napoletano2019}  in the previous section.}}

\subsection{Node Embedding Techniques}
\label{sec:sample2_net}

 {\color{black}
In GRL, an embedding is a graph representation obtained through a function that preserves some properties of $G$ while projecting its elements into a lower-dimensional space \citep{Goyal:04}. There are two approaches to defining embeddings. One involves representing the entire graph in a lower-dimensional vector space, while the other focuses on representing a part of the graph in the same space (node/vertex, edge, substructure). The latter is of particular interest in this paper, especially the node granularity. The embeddings can be used as input attributes for downstream tasks. In this study, following the generation of embeddings, the subsequent task is to employ these embeddings to guide the GRASP heuristic in solving the C-IGDP. The main objective is to arrange the vertices in their respective layers while adhering to predefined constraints, aiming to reduce the number of arc crossings in this graph.

Embedding generation is challenging, with several techniques and taxonomies proposed. Following \cite{cai:03}, this study focused on two main categories: matrix factorization and deep learning.

{\color{black}Matrix factorization is an embedding model where a matrix represents the graph properties, and its factorization describes the embeddings. Deep learning approaches can employ random walk strategies and an autoencoder model. On the one hand, the random walk approaches are stochastic strategies designed to capture the local structure of graphs, generating training samples. The models are trained to predict neighboring node occurrences, adjusting embeddings to maximize this probability and capture graph relationships. On the other hand, the autoencoder comprises the encoder and decoder to acquire an efficient data representation through compression and subsequent accurate reconstruction \citep{cai:03}. The encoder maps the graph nodes to low-dimensional vectors (embeddings). The decoder subsequently employs these embeddings to reconstruct information regarding the nodes' neighborhood in the original graph. Encoder-decoder architectures are machine learning models trained to minimize information loss during the reconstruction process. The standard typically employs pairwise decoders to predict similarity between nodes, with the loss function serving as a measure of the discrepancy between predicted and actual values derived from a predefined similarity measure. Methods for embedding generation may differ depending on the choice of decoder function, similarity measure, and loss function \citep{Hamiltonbook}. }

Next, we briefly describe the graph embedding strategies used in the proposed learning-based metaheuristics. 

\begin{itemize}
\item \textit{High-Order Proximity preserved Embedding} (\textit{HOPE}) \citep{Mingdong} considers the asymmetric transitivity properties of the graph to preserve high-order proximity between nodes. The decoder is based on the inner product between nodes, assumed to be proportional to their similarity. The similarity measure is based on the Katz index. This index counts paths of all lengths between pairs of nodes, offering a comprehensive view of relationships in the graph. The loss function consists of the mean squared error between the decoder and the similarity matrix, ensuring proper optimization of the embeddings.

    \item \textit{Node2Vec} \citep{Grover}  employs random walks on the graph to generate embeddings, adjusting the transition probability between neighboring nodes, where parameters control the bias of these walks. These walks produce node vector representations, following the concept of the skip-gram model from the so-called word2vec adapted for graphs. The algorithm operates in three main stages: preprocessing to compute transition probabilities, simulation of random walks, and optimization of the vector representations. The encoder corresponds to the preprocessing and random walk simulation stages. The decoder is the optimization stage, where the vector representations are adjusted to reconstruct the graph's structure. The decoder is proportional to the similarity function that, in this case, calculates the probability of visiting node $v$ in a fixed-length random walk starting at a node $u$. This measure reflects the frequency with which node $v$ is reached from $u$ in random walks of this length, providing information about the proximity or connection between these nodes in the graph. Ultimately, the vector representations learned by Node2Vec are refined by maximizing the probability of reconstructing the original graph through the minimization of a cross-entropy loss function.

    \item  \textit{Structural Deep Network Embedding} (\textit{SDNE}) employs deep autoencoders to generate graph embeddings, preserving both first-order proximity (direct node neighborhood) and second-order proximity (relationships between node neighbors). SDNE uses an unsupervised approach to efficiently reconstruct node neighborhoods in the graph and a supervised part, using Laplacian Eigenmaps, to penalize cases where similar nodes are mapped far apart from each other in the embedding space \citep{Goyal:04}. In summary, SDNE harnesses the capability of deep autoencoders to capture nonlinearities in data, generating embeddings that preserve the proximity structures of the graph, making it an effective technique for graph analysis and learning.

 \item  \textit{Spectral} \citep{Tang} is a classical graph learning algorithm based on spectral clustering. Spectral is grounded on the eigen-de\-com\-po\-si\-tion of the normalized Laplacian matrix of the given graph. The matrix of the eigenvectors associated with the smallest eigenvalues of the normalized Laplacian matrix is used to define the node embeddings of the graph.
\end{itemize}
}
\textcolor{black}{The embeddings generated through these techniques are used to guide the  construction phase of the proposed GL-GRASP heuristics to the C-IGDP. Specifically, they serve as structural references for node placement, enriching the classical layout heuristics with global graph information, as discussed in the next section.}

\subsection{\textcolor{black}{GL-based Construction Phase}}
\textcolor{black}{The construction phase of the proposed method builds upon the classical barycenter heuristic, which places a new node at the average position of its neighbors to reduce edge crossings. While effective, the barycenter heuristic relies solely on local geometric information, which can lead to an unbalanced use of space in the final layout. To address this limitation, we incorporate insights from graph embedding strategies to capture global structural relationships. In the proposed strategy, node placement is guided by the Euclidean distance in the embedding space, allowing positions to reflect both local and global graph properties. Incremental nodes are placed using one of three randomly selected strategies: near the closest neighbor (to preserve community structure), near the most distant neighbor (to avoid clustering and promote spatial balance), or at an intermediate point (to facilitate smooth transitions and prevent both over-concentration and excessive dispersion). These options enhance the exploratory capabilities of the method and reduce the rigidity of neighborhood-based placement. Overall, the approach aims to minimize edge crossings while maintaining the structural coherence captured by the node embeddings.} 

The construction phase of the learning-based GRASP heuristics proposed in this paper employs the node embeddings generated by the strategies described in the previous section. For this, the data preprocessing stage occurs, described next.

\vspace{0.5cm}
\hrule
\vspace{0.05cm}
\noindent\textit{Node Embedding-based Arc Attributes} 
\vspace{0.05cm}
\hrule
\begin{enumerate}[\textit{Step} 1]
\item A literature-based embedding technique is chosen to generate embeddings for all vertices of $IG$. 
\item The embeddings are transformed and projected onto a plane, yielding Cartesian coordinates (xy-coordinates), considering the Principal Component Analysis (PCA) \textcolor{black}{\citep{pca}}.
\item The pairwise Euclidean distance between adjacent vertices $u , v \in V$, referred to as $d(u,v)$, is computed considering their Cartesian coordinates, and stored as arc attributes.
\end{enumerate}
\hrule
\vspace{0.5cm}

The introduced construction method starts with an initial partial solution (initial layout of the given graph) and iteratively adds the incremental vertices to the partial solution until all incremental vertices have been added to the solution. In this context, $\mathcal{I}'$ is said a partial solution (a partial incremental drawing) if $\mathcal{I}'$ meets the conditions imposed by Claims~\ref{relativeprecedence} and \ref{absoluteprecedende}, when considering the subgraph of $IG$ spanned by the original and incremental vertices already in the partial solution $\mathcal{I}'$. The following steps describes the construction phase.

\vspace{0.5cm}
\hrule
\vspace{0.05cm}
\noindent\textit{Construction Phase} 
\vspace{0.05cm}
\hrule
\begin{enumerate}[\textit{Step} 1]
\item Let $\mathcal{I}'$ be the initial drawing, where only the original vertices belong to it.

\item Let the initial candidate list $CL$ be $\mathop{\hat{V}}$.
\item Let $\mathcal{G} (u)$ be the minimum distance between $u$ and its neighbors already in the partial solution. Considering all the computed distances between a vertex $u$ and its neighbors, $\mathcal{G} (u)$ can be defined as in Equation~\eqref{gu}.\label{step3}
\begin{equation}
    \mathcal{G} (u) =  \mathop{\min_{(u,v),(v,u)\in \mathcal{I}'}d(v,u)}\label{gu}
\end{equation}
\item Consider $RCL = \{ \mathop{\nu \in CL:} \mathcal{G} \mathop{(\nu)}  \mathop{\leq \xi} \}$, where $\xi$ is calculated in Equation~\eqref{threshold}.

\begin{equation}
\mathop{\xi = \min_{\nu \in CL}} \mathcal{G} \mathop{(\nu) + \varphi ( \max_{\nu \in CL}}  \mathcal{G} \mathop{(\nu) - \min_{\nu \in CL}}  \mathcal{G} \mathop{(\nu))} 
\label{threshold}
\end{equation}

Parameter $\varphi$ is randomly chosen within the interval $[0,1]$, balancing diversity and quality in the search \citep{Sanchez2017}.

\item Select a vertex \( v \in \text{RCL} \) at random \textcolor{black}{and construct the set of possible positions \( P \), initially empty}:\label{step5}
\begin{itemize}
    \item[(a)] \textcolor{black}{For each neighboring layer \( \lambda' \) that contains a node adjacent to \( v \), randomly select one of the two possible positions to add to \( P \):}
    \begin{itemize}
        \item[(i)] The closest available position to the \textit{nearest} neighbor of \( v \) in \( \lambda' \);
        \item[(ii)] The closest available position to the \textit{farthest} neighbor of \( v \) in \( \lambda' \).
    \end{itemize}
    
    \item[(b)] \textcolor{black}{Add to \( P \)} the closest available position to the \textit{average} of the positions already in \( P \).
    
    \item[(c)] \textcolor{black}{Finally, arbitrarily choose one of the positions in \( P \) to place node \( v \) in \( I' \).}
\end{itemize}

\item Remove $v$ from $CL$ and if $CL$ is not empty go to \textit{Step} \ref{step3}.
\item Return $\mathcal{I}'$.
\end{enumerate}
\hrule

\vspace{0.5cm}

Some optimizations are possible to avoid redundant operations in updating the $RCL$. When adding a vertex $u\in RCL$ to the partial solution, only vertices adjacent to $u$, let us say $v$,  are re-evaluated to update $\mathcal{G} (v)$. Therefore, it is not necessary to re-compute all minimum distance values if \textit{Step}~\ref{step3} has been performed at least once. Moreover, the random choice of the position where the vertex will be placed in the drawing allows significant diversity for the method in preliminary experiments. In this process, we guarantee that the candidate positions are close to neighbor vertices.

After the construction, the local search phase is applied to the solution obtained in the first phase. The method is similar to the one proposed by \cite{napoletano2019}. It follows the best improvement choice and first performs all possible swap movements. When it is not possible to further improve the solution through this movement, the insert movement starts and halts if no better solution is found. 

\begin{figure}[!h]
\includegraphics[scale=1]{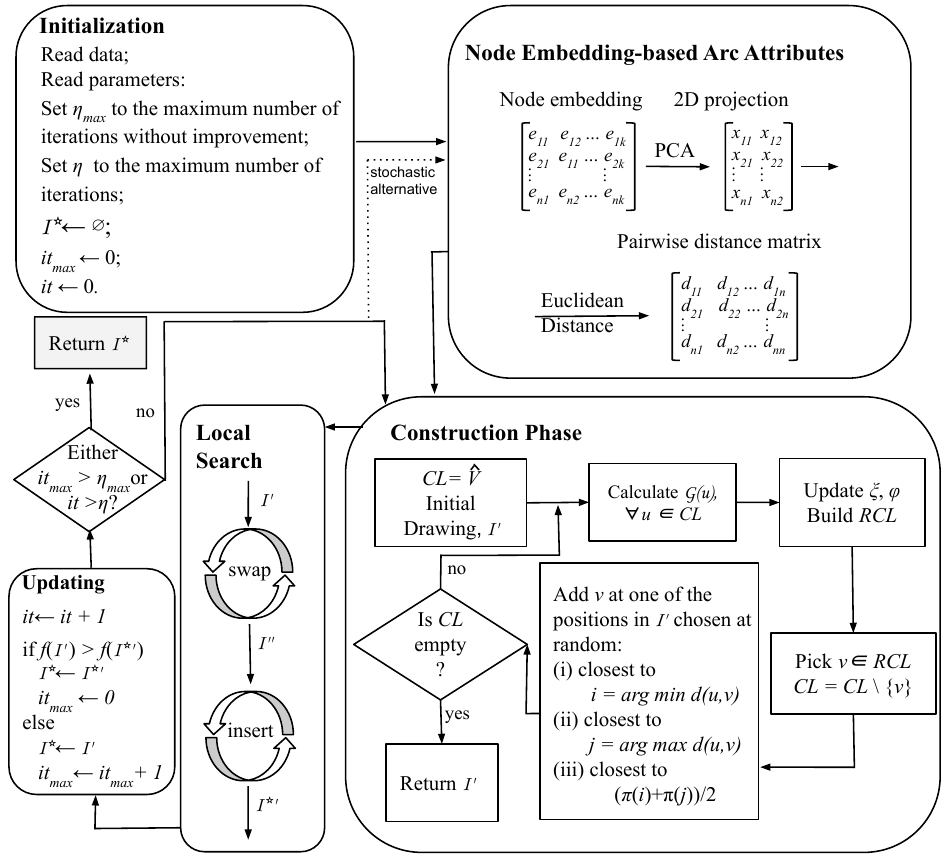}
\caption{Flowchart of the Graph Learning GRASP (GL-GRASP) introduced in this paper.} \label{flowchart}
\end{figure}

The maximum number of the GL-GRASP iterations is defined \textit{a priori}, here identified by $\eta$. Moreover, to avoid unnecessary GRASP iterations, if no better solutions are found in the last $\eta_{max}$ iterations, the algorithm stops and returns the best overall solution. A flowchart summarizing the learning-based method is presented in Figure~\ref{flowchart}. In this figure, the following notation simplifications were assumed, without loss of generality.
\begin{enumerate}[(i)]
    \item The vertex labels are integer number from 1 to $n$, the total number of vertices from the incremental graph;
    \item The notation $d_{ij}$ means $d(i,j)$;
    \item We omit the $\lambda$ index in $\pi$ since its value is not important for the equation, only the position of the vertices in the corresponding layer.
\end{enumerate}

{\color{black}In the \textit{Initialization} block in Figure~\ref{flowchart},  the input data and parameters of the GL-GRASP are read and/or initialized. The parameters $\eta$ and $\eta_{\max}$ are initialized. The variable $I^*$, initialized as empty, aims to store the best solution found during the process. Additionally, the counters $it_{\max}$ and $it$ start at zero, representing, respectively, the number of iterations without improvement and the total number of iterations.

In the \textit{Node Embedding-based Arc Attributes} block, the extraction and preprocessing of the node embedding matrix occurs, as described earlier in this section. The dimension of the node-embedding matrix is \( n \times k \), where \( n \) represents the total number of nodes in the incremental graph and \( k \) is the embedding dimension. This matrix stores an embedding vector for each node, which is projected onto a two-dimensional space using the PCA technique. Each row of the 2D projection matrix corresponds to an ordered pair of Cartesian coordinates of the corresponding node. The Euclidean distance is then computed between each pair of adjacent nodes, producing the pairwise distance matrix. Matrix elements corresponding to non-adjacent node pairs are marked with the character 'NA'.

The \textit{Construction Phase} block illustrates all the steps for constructing the initial solution $I'$, as discussed earlier in this section. After the solution is complete, it goes through the local search (\textit{Local Search} block) that consists of the swap and insert movements. The counters are incremented, and the best solution obtained in the GL-GRASP iteration is updated if necessary. A new solution is constructed if none of the stopping criteria is met: $it_{\max} > \eta_{\max}$, indicating that the limit of iterations without updating the best solution has been exceeded, or $it > \eta$, meaning that the total number of allowed iterations has been reached. 

One may notice that there are two possible strategies for constructing the solution in the next iteration. The first is a stochastic approach, which recalculates the node embeddings and the distance matrix before constructing the new solution. The second uses the previously computed distance matrix to generate the solution for the next iteration of GL-GRASP. The aim of the former is to increase the diversity of GL-GRASP by exploring different embeddings produced by stochastic node embedding algorithms. However, since not all node embedding strategies are stochastic, this alternative is only applicable when the algorithm is non-deterministic.}
 
 In this paper, four node embedding techniques composed the node embedding-based arc attributes of the incremental graphs to guide the construction phase of GRASP: \textit{Spectral} (SPEC), HOPE, \textit{Node2Vec} (N2V), and SDNE.  Therefore, for each of these node embedding algorithms, a version of the GL-GRASP is developed. As N2V and SDNE are stochastic, we also considered the version in which the ``stochastic alternative'' is chosen in every GRASP iteration. Therefore, the total of GL-GRASP heuristics tested in this paper is six. 



\section{Computational Experiments}
\label{sec:sample4}

This section presents the computational experiments performed to assess the learning-based GRASP heuristics. We ran the experiments in a sequential processing system comprising four dedicated high-memory nodes. Each node has two Intel Xeon E5-2667v4 processors of 3.2 GHz, 8 cores, and 512 GB DDR3 1866 MHz memory. This setup resulted in a cumulative configuration of 64 processing cores and 2 TB of memory.

In summary, the following GL-GRASP heuristics were tested, considering the different node embedding strategies and variants of N2V and SDNE:
\begin{itemize}
\item G\_SPEC: GL-GRASP heuristic using SPEC to extract the node embeddings;
\item G\_HOPE: GL-GRASP heuristic using HOPE to extract the node embeddings;
\item G\_N2V: GL-GRASP heuristic using N2V to extract the node embeddings;
\item G\_SDNE: GL-GRASP heuristic using SDNE to extract the node embeddings;
\item G\_N2V*: GL-GRASP heuristic using stochastic N2V to extract the node embeddings;
\item G\_SDNE*: GL-GRASP heuristic using stochastic SDNE to extract the node embeddings;
\end{itemize}

 The employed implementation of the node embedding strategies was found in the CogDL library \citep{cen2021cogdl}. {\color{black} Besides, for a comparative analysis we implemented the literature state-of-the-art methods, the GRASP heuristics introduced by \cite{napoletano2019}, referred  to as GRASP2 ($C_2$ + local search) and GRASP3 ($C_3$ + local search). The local search of all methods, the GL-GRASP heuristics and the literature GRASP heuristics, are the same. All algorithms have been implemented in the Python programming language and codes are available at \url{https://github.com/bcbraga/CIGDP-DL.git}.} 
 
{\color{black}This section presents three experiments. The first compares the performance of the GL-GRASP heuristics with one another. The second evaluates the most promising GL-GRASP variants against the state-of-the-art GRASP heuristics proposed by \citet{napoletano2019}, using benchmark instances. \textcolor{black}{To assess the quality of the heuristic methods, in addition to the comparison with heuristics from the literature, we ran the experiments using the Gurobi solver (version 10.0.2) \citep{gurobi}, which was able to find the optimal solution for all benchmark instances within a reasonable time. The results, along with the corresponding execution times, are provided in the Supplementary Material.} To further evaluate the heuristics on more challenging instances, a third experiment was conducted using larger instances for which Gurobi was unable to prove optimality or find an upper bound within a reasonable time.}

The next section presents a brief description of the employed datasets.

\subsection{Instances}
{\color{black}This section briefly describes the instances used in the three experiments. Benchmark datasets were employed in the first and second experiments, whereas newly generated instances were used in the third experiment. Further details on these instances are provided in the respective sections.}

\subsubsection{Benchmark Data}

The benchmark dataset comprises 609 test cases derived from the same 240 instances used by \cite{napoletano2019}, which are available at \textcolor{black}{\url{https://github.com/bcbraga/CIGDP-DL/tree/main/data}}. These instances represent hierarchical graphs with 5 to 30 non-incremental vertices per layer. The number of layers of these graphs is 2, 6, 13, and 20 layers, where the graph densities (average number of arcs per vertex) are 0.065, 0.175, and 0.300. For each combination of quantity of layers and density, we generated 20 instances, totaling 240 instances. The amount of incremental vertices of 120 instances is $20 \%$  of the number of original vertices  (Inc = 0.2). The number of incremental vertices of the other 120 instances is  $60 \%$ of non-incremental vertices (Inc = 0.6). For each of the 240 instances, the parameter $d$ received the values 1, 2 and 3, following the literature methodology for generating instances. It is important to highlight that the size of the set of incremental vertices in each layer must be greater than or equal to \(d\). Considering this criterion, 111 out of the 720 instances did not meet the requirement imposed by the authors of a minimum number of incremental vertices per layer in the computational experiments. Therefore, the dataset contains a total of 609 test cases.

{\color{black}\subsubsection{Newly Generated Instances}

{\color{black}This set of instances was generated following the same structural principles used to create the benchmark instances. They are publicly available at: \url{https://github.com/bcbraga/CIGDP-DL/tree/main/data_extra}.

We generated one instance for each of the following numbers of layers: $\Lambda \in \{2, 3, 4, 5\}$. All instances have a graph density of 0.5 ($\rho = 0.05$) and a node incremental ratio of 0.60 ($Inc = 0.60$). For each instance, the parameter $d$ considered in the experiments takes values in $\{1, 2, 3\}$. Additionally, the number of non-incremental nodes per layer was randomly selected within the range of 60 to 80. The degree of the incremental nodes ranges from 1\% to 10\% of the number of nodes in their neighboring layers. 
Table \ref{tab:incgraph-characteristics} provides an overview of the newly generated instances, presenting the number of layers, the number of incremental and original nodes per layer, and the number of incremental and original arcs leaving each layer.} 

\begin{table}[htp!]
\footnotesize
\centering
\color{black}
\caption{\textcolor{black}{Characteristics of the newly generated graphs.}}
\label{tab:incgraph-characteristics}
\begin{tabular}{ccccccc} 
\toprule
\textbf{Instance} & $\Lambda$ &\textbf{$\lambda$} & \textbf{$|\widehat{V}^{\lambda}|$} & \textbf{$|V^{\lambda}|$} &$|\widehat{A}^{\lambda}|$  & $|A^{\lambda}|$ \\
\midrule
\multirow{2}{*}{incgraph\_2} &\multirow{2}{*}{2}
& 1 & 38 & 64 & 319 & 2110 \\
& & 2 & 36 & 61 & --- & --- \\
\midrule
\multirow{3}{*}{incgraph\_3} &\multirow{3}{*}{3}
& 1 & 44 & 74 & 308 & 2556 \\
& & 2 & 41 & 69 & 299 & 2228 \\
& & 3 & 36 & 60 & --- & --- \\
\midrule
\multirow{4}{*}{incgraph\_4} &\multirow{4}{*}{4}
& 1 & 46 & 77 & 289 & 2776 \\
& & 2 & 43 & 72 & 230 & 2233 \\
& & 3 & 37 & 62 & 268 & 2260 \\
& & 4 & 40 & 68 & --- & --- \\
\midrule
\multirow{5}{*}{incgraph\_5} &\multirow{5}{*}{5}
& 1 & 40 & 68 & 210 & 2143 \\
& & 2 & 37 & 63 & 265 & 2174 \\
& & 3 & 41 & 69 & 225 & 2278 \\
& & 4 & 39 & 66 & 337 & 2472 \\
& & 5 & 42 & 70 & --- & --- \\
\bottomrule
\end{tabular}
\end{table}

}

\subsection{Parameter Setup}

As described in Figure~\ref{flowchart}, we considered the Euclidean two-dimensional distance to generate the pairwise distance matrix from the obtained node embeddings, a choice grounded in the literature due to its effectiveness and simplicity. \textcolor{black}{Given the complexities involved in selecting parameters for node embedding methods, we utilized the default settings from the CogDL library \citep{cen2021cogdl}. These defaults have been empirically fine-tuned across diverse graph structures, providing robust and practical starting points for model training. While optimizing these baseline parameters for our specific dataset was not the focus of this study, relying on these validated CogDL defaults is a reasonable and usual approach. Detailed information on the default parameters for each embedding method (HOPE, SPEC, N2V, and SDNE), including their values and functional descriptions, can be found in the Supplementary Material.} 

It is worth emphasizing that, in the HOPE and SPEC methods, the embedding dimension (parameter \( k \) in Figure~\ref{flowchart}) was adjusted based on the number of vertices (\( n \)) in the input graph, while the default value suggested in the literature, \( k = 128 \), was used for the other methods.
According to the CogDL documentation, the $k$ value depends on the number of computed singular values $sv$ and the number of vertices of the graph. Considering the total number of vertices \(n\), the values of \(k\) were defined as follows: for the HOPE algorithm, \(k = 2n - 1\), since \(sv = k/2\), reflecting the utilization of both the left and right singular vectors of the decomposition matrix in the implementation. For the SPEC algorithm, \(k = n - 1\), with \(sv = k\), indicating the use of only the left singular vectors of the decomposition matrix.

Regarding the parameter $\mathop{\varphi}$ of GRASP, we followed the approach proposed by \cite{ResendeRibeiro} to meet the difficulty of adjusting this parameter deterministically. In this regard, we adhere to the common practice in the literature, where a value of $\mathop{\varphi}$ is randomly chosen at each iteration of GRASP, providing a pragmatic solution to this issue. It is worth noting that the authors in \cite{napoletano2019} fine-tuned this parameter. Their analysis demonstrated that the optimal choice for this parameter would be random values. The GRASP parameters $\eta$ and $\eta_{max}$ were set to 100 and 20, respectively, achieved after preliminary experiments evaluating the trade-off between time and solution quality. These parameter values were employed not only for the GL-GRASP heuristics but also for the state-of-the-art GRASP heuristics GRASP2 and GRASP3. 

\subsection{Assessment Metrics}

To assess the performance of the tested heuristics, we considered several measures and strategies. To define them, consider first the following measures that correspond to performance values of the algorithm calculated per instance for each independent execution.
\begin{itemize}

\item time: execution time in seconds (\textcolor{black}{including the embedding learning time in the GL\_GRASP heuristics});
\item gap: the difference between the heuristic and the optimal solutions divided by the optimal solution;
\item NPI: Normalized Primal Integral, the primal integral scoring system proposed by \cite{Timo} to evaluate each independent execution of an algorithm considering both solution quality and execution time. This measure evaluates the method's performance over time by assessing the convergence towards the final solution by considering the progress of the primal solutions. The numerical estimation of the NPI is explained next.

\end{itemize}

Let $f_{heur}^*$ be the best solution value among all GL-GRASP heuristics for a specific instance problem, and $f(0)=1.1\times f^*_{heur}$. Moreover, consider $T_{max}$ the maximum time (in seconds) among all GL-GRASP heuristics for a given instance. For each\ independent execution of a GL-GRASP heuristic, we keep a sequence of the $n_s$ solutions better than $f(0)$ found in the GL-GRASP iterations sorted in decreasing order of time to reach the given solution. The solution value of the $i$-th solution from the sequence is referred to as $f(i)$ and the corresponding time $t(i)$. Consider $t(0)=0$. The Normalized Primal Integral (NPI) is then calculated according to Equation \eqref{eq:1npi}.

\begin{equation}
\mathop{\mbox{NPI} = \cfrac{\sum_{i=1}^{n_s}[f(i-1) \times (t(i) - t(i-1))] + f(n_s) \times (T_{max} - t(n_s))}{T_{max} \times f_{heur}^*}} 
\label{eq:1npi}
\end{equation}

The metrics reported in the experiments provide the \textcolor{black}{average}  of the algorithms over a set of instances in a number of independent executions. 

To better illustrate the values of some metrics in the instance set, we present their performance profiles. \cite{Dolan2002} proposed the performance profiles, which are graphics relating the proportion of instance problems for which an algorithm achieved at least the performance indicated by the x-axis. For computing the performance measure, one must consider the best overall result multiplied by a factor, starting from 1 up to the required factor value to achieve the worst overall result for that instance. The x-axis corresponds to the factor values. Therefore, if the factor is 1.05, the y-axis of the given curve indicates the proportion of instance problems that the corresponding algorithm achieved a solution better than or equal to 1.05 multiplied by the best overall solution value. The curve of an algorithm $A$ dominating an algorithm $B$ means that $A$ outperforms $B$. These profiles are used in this paper to evaluate the behavior of the results achieved by the algorithms for a better evaluation of the performance of the algorithms.

\subsection{First Experiment}

This experiment contrasts the proposed GL-GRASP heuristics in benchmark instances.  We ran each heuristic  30 times for each tested instance and reported the average results. Tables \ref{table:averageresults} and  \ref{table:averageresults_npi} show the average times, gaps and NPIs of the methods, dividing the instances into classes according to the number of layers ($\Lambda$) and density values ($\rho$). We highlight in bold the best results per group of instances.

\begin{table}[!htp]
\caption{Table presenting the \textcolor{black}{average} gaps and times  by groups of instances with 2, 6, 13 and 20 layers and the three considered density values.}\label{table:averageresults}
\begin{adjustbox}{max width=1.0\textwidth}
\begin{tabular}{ll|ll|ll|ll|ll|ll|ll}
\hline
\multicolumn{2}{l|}{Instances}                    & \multicolumn{2}{l|}{G\_N2V}     & \multicolumn{2}{l|}{G\_N2V*}    & \multicolumn{2}{l|}{G\_HOPE}    & \multicolumn{2}{l|}{G\_SDNE}    & \multicolumn{2}{l|}{G\_SDNE*}   & \multicolumn{2}{l}{G\_SPEC}    \\ \hline
\multicolumn{1}{l|}{$\Lambda$}              & $\rho$ & \multicolumn{1}{l|}{\textcolor{black}{gap}} & \textcolor{black}{time} & \multicolumn{1}{l|}{\textcolor{black}{gap}} & \textcolor{black}{time} & \multicolumn{1}{l|}{\textcolor{black}{gap}} & \textcolor{black}{time} & \multicolumn{1}{l|}{\textcolor{black}{gap}} & \textcolor{black}{time} & \multicolumn{1}{l|}{\textcolor{black}{gap}} & \textcolor{black}{time} & \multicolumn{1}{l|}{\textcolor{black}{gap}} & \textcolor{black}{time} \\ \hline
\multicolumn{1}{c|}{\multirow{3}{*}{2}}  & 0.065   & \multicolumn{1}{l|}{0.311}    &18.49      & \multicolumn{1}{l|}{0.283}    &53.82      & \multicolumn{1}{l|}{0.252}    &\textbf{14.31}      & \multicolumn{1}{l|}{0.244}    &20.17      & \multicolumn{1}{l|}{\textbf{0.243}}    &96.06 & \multicolumn{1}{l|}{0.252}    &15.91      \\ 
\multicolumn{1}{c|}{}                    & 0.175   & \multicolumn{1}{l|}{0.229}    &50.56      & \multicolumn{1}{l|}{0.165}    &92.23      & \multicolumn{1}{l|}{\textbf{0.089}}    &40.95      & \multicolumn{1}{l|}{0.175}    &48.40      & \multicolumn{1}{l|}{0.179}    &130.76      & \multicolumn{1}{l|}{0.141}    &\textbf{45.92}      \\
\multicolumn{1}{c|}{}                    & 0.300   & \multicolumn{1}{l|}{0.188}    &102.46      & \multicolumn{1}{l|}{0.177}    &141.93      & \multicolumn{1}{l|}{0.183}    &\textbf{86.21}      & \multicolumn{1}{l|}{0.130}    &90.93      & \multicolumn{1}{l|}{0.134}    &190.20      & \multicolumn{1}{l|}{\textbf{0.119}}    &97.17     \\ \hline
\multicolumn{1}{c|}{\multirow{3}{*}{6}}  & 0.065   & \multicolumn{1}{l|}{0.468}    &170.45      & \multicolumn{1}{l|}{0.469}    &356.78      & \multicolumn{1}{l|}{0.434}    &\textbf{142.16}      & \multicolumn{1}{l|}{0.441}    &152.74      & \multicolumn{1}{l|}{0.446}    &408.34      & \multicolumn{1}{l|}{\textbf{0.416}}    &151.61      \\ 
\multicolumn{1}{l|}{}                    & 0.175   & \multicolumn{1}{l|}{0.170}    &378.87      & \multicolumn{1}{l|}{\textbf{0.149}}    &549.19      & \multicolumn{1}{l|}{0.157}    &328.65      & \multicolumn{1}{l|}{0.170}    &\textbf{328.59}      & \multicolumn{1}{l|}{0.179}    &579.68      & \multicolumn{1}{l|}{0.155}    &356.94      \\
\multicolumn{1}{l|}{}                    & 0.300   & \multicolumn{1}{l|}{0.065}    &809.82      & \multicolumn{1}{l|}{\textbf{0.056}}  &  1003.13     & \multicolumn{1}{l|}{0.066}    &747.84      & \multicolumn{1}{l|}{0.063}    &\textbf{707.79 }     & \multicolumn{1}{l|}{0.060}    &1049.08      & \multicolumn{1}{l|}{0.080}    &797.98      \\ \hline
\multicolumn{1}{c|}{\multirow{3}{*}{13}} & 0.065   & \multicolumn{1}{l|}{0.707}    &620.44      & \multicolumn{1}{l|}{0.692}    &1150.74      & \multicolumn{1}{l|}{0.703}    &549.20      & \multicolumn{1}{l|}{0.698}    &\textbf{527.33 }     & \multicolumn{1}{l|}{\textbf{0.691}}    &1280.02      & \multicolumn{1}{l|}{0.705}    &587.83      \\
\multicolumn{1}{l|}{}                    & 0.175   & \multicolumn{1}{l|}{0.247}    &1461.46      & \multicolumn{1}{l|}{0.249}    &1826.33      & \multicolumn{1}{l|}{0.235}    &1224.43      & \multicolumn{1}{l|}{0.234}    &\textbf{1148.91 }     & \multicolumn{1}{l|}{\textbf{0.233}}    &2030.60      & \multicolumn{1}{l|}{\textbf{0.233}}    &1279.54      \\
\multicolumn{1}{l|}{}                    & 0.300   & \multicolumn{1}{l|}{0.118}    &2744.94      & \multicolumn{1}{l|}{0.114}    &3126.39      & \multicolumn{1}{l|}{0.118}    &2483.97      & \multicolumn{1}{l|}{0.118}    &\textbf{2337.73 }     & \multicolumn{1}{l|}{\textbf{0.110}}    &3349.63     & \multicolumn{1}{l|}{0.118}    &2649.88      \\ \hline
\multicolumn{1}{c|}{\multirow{3}{*}{20}} & 0.065   & \multicolumn{1}{l|}{0.723}    &1236.86      & \multicolumn{1}{l|}{\textbf{0.722}}    &2069.93      & \multicolumn{1}{l|}{0.728}    &1166.22      & \multicolumn{1}{l|}{0.734}    &\textbf{1153.75}      & \multicolumn{1}{l|}{0.730}    &2549.18      & \multicolumn{1}{l|}{0.730}    &1220.78      \\
\multicolumn{1}{l|}{}                    & 0.175   & \multicolumn{1}{l|}{0.282}    &2740.13      & \multicolumn{1}{l|}{0.279}    &3423.48      & \multicolumn{1}{l|}{0.281}    &2475.73      & \multicolumn{1}{l|}{0.276}    &\textbf{2337.61}      & \multicolumn{1}{l|}{0.275}    &3810.23      & \multicolumn{1}{l|}{\textbf{0.271}}    &2616.52      \\
\multicolumn{1}{l|}{}                    & 0.300   & \multicolumn{1}{l|}{0.117}    &5608.73      & \multicolumn{1}{l|}{\textbf{0.114}}    &6090.51    & \multicolumn{1}{l|}{0.115}    &\textbf{4823.11}      & \multicolumn{1}{l|}{0.117}    &4984.82      & \multicolumn{1}{l|}{0.115}    &6871.25      & \multicolumn{1}{l|}{\textbf{0.114}}    &5284.52      \\ \hline

\multicolumn{2}{l|}{\textbf{\textcolor{black}{Mean}}}  
& \multicolumn{1}{l|}{\textcolor{black}{0.302}} 
& \text{\textcolor{black}{1328.60}} 
& \multicolumn{1}{l|}{\textcolor{black}{0.289}} 
& \text{\textcolor{black}{1657.04}} 
& \multicolumn{1}{l|}{\textcolor{black}{0.280}} 
& \text{\textcolor{black}{1173.57}} 
& \multicolumn{1}{l|}{\textcolor{black}{0.283}} 
& \text{\textcolor{black}{1153.23}} 
& \multicolumn{1}{l|}{\textcolor{black}{0.283}} 
& \text{\textcolor{black}{1862.09}} 
& \multicolumn{1}{l|}{\textcolor{black}{\textbf{0.278}}} 
& \text{\textcolor{black}{1258.72}} \\ \hline

\end{tabular}
\end{adjustbox}
\end{table}

According to Table~\ref{table:averageresults}, G\_SPEC was the algorithm with a higher number of best average haps, being the top in 5 classes of instances. Moreover, G\_SPEC had the second-best gaps in 4 out of the 7 remaining classes.  G\_N2V* and  G\_SDNE* achieved the best gaps in 4 groups of instances each. G\_HOPE obtained the best \textcolor{black}{average gap} in one class of instances, whereas the other methods have not achieved the best gaps in any class. Nevertheless, the G\_SDNE* and G\_N2V* computational times were significantly higher than the non-stochastic alternative frameworks. Despite not being as time-consuming as G\_SDNE* and G\_N2V*, the computational times of G\_SPEC were not as competitive as those required by the methods based on deep learning. In line with this, one can observe that G\_SDNE had a good performance considering both the gap and computational time. It had the best average times in 7 classes and, despite not achieving the best average gaps in any class of instances, its \textcolor{black}{average gap} values were very competitive with the top best. G\_HOPE also stood out, presenting the second best in 4 groups of instances besides being the best in 1 class. G\_N2V was the method with the least competitiveness in the gap and computational time. {\color{black} The mean results show that G\_SPEC achieved the lowest mean gap, while G\_SDNE was the fastest method. Considering both average gap and time, G\_HOPE obtained the second-best overall performance.}

As discussed earlier, the NPI is a primal integral measure that evaluates the evolution of primal heuristics along the execution time. According to \cite{Timo}, this measure captures the overall solving process more adequately. Our goal is to investigate graph learning in helping the search process of a state-of-the-art metaheuristic by using information on the latent structure of the graphs. In line with this, the NPI can help us better infer the behavior of the methods along the iterations.
Table~\ref{table:averageresults_npi} reports the \textcolor{black}{average NPI and time} values of the classes of instances divided into groups as in Table~\ref{table:averageresults}.

\begin{table}[!htp]
\caption{Table presenting the \textcolor{black}{average NPI and times} per group of instances with 2, 6, 13 and 20 layers and the three considered density values.}\label{table:averageresults_npi}
\begin{adjustbox}{max width=1.\textwidth}
\begin{tabular}{ll|ll|ll|ll|ll|ll|ll}
\hline
\multicolumn{2}{l|}{Instances}                    & \multicolumn{2}{l|}{G\_N2V}     & \multicolumn{2}{l|}{G\_N2V*}    & \multicolumn{2}{l|}{G\_HOPE}    & \multicolumn{2}{l|}{G\_SDNE}    & \multicolumn{2}{l|}{G\_SDNE*}   & \multicolumn{2}{l}{G\_SPEC}    \\ \hline
\multicolumn{1}{l|}{$\Lambda$}              & $\rho$ & \multicolumn{1}{l|}{\textcolor{black}{NPI}} & \textcolor{black}{time} & \multicolumn{1}{l|}{\textcolor{black}{NPI}} & \textcolor{black}{time} & \multicolumn{1}{l|}{\textcolor{black}{NPI}} &  \textcolor{black}{time} & \multicolumn{1}{l|}{\textcolor{black}{NPI}} &  \textcolor{black}{time} & \multicolumn{1}{l|}{\textcolor{black}{NPI}} &  \textcolor{black}{time} & \multicolumn{1}{l|}{\textcolor{black}{NPI}} &  \textcolor{black}{time} \\ \hline
\multicolumn{1}{c|}{\multirow{3}{*}{2}}  & 0.065   & \multicolumn{1}{l|}{0.207}    &18.49      & \multicolumn{1}{l|}{0.335}    &53.82      & \multicolumn{1}{l|}{0.136}    &\textbf{14.31}      & \multicolumn{1}{l|}{\textbf{0.130}}    &20.17      & \multicolumn{1}{l|}{0.469}    &96.06 & \multicolumn{1}{l|}{0.142}    &15.91      \\ 
\multicolumn{1}{c|}{}                    & 0.175   & \multicolumn{1}{l|}{0.282}    &50.56      & \multicolumn{1}{l|}{0.359}    &92.23      & \multicolumn{1}{l|}{\textbf{0.131}}    &40.95      & \multicolumn{1}{l|}{0.215}    &48.40      & \multicolumn{1}{l|}{0.516}    &130.76      & \multicolumn{1}{l|}{0.190}    &\textbf{45.92}      \\
\multicolumn{1}{c|}{}                    & 0.300   & \multicolumn{1}{l|}{0.245}    &102.46      & \multicolumn{1}{l|}{0.346}    &141.93      & \multicolumn{1}{l|}{0.221}    &\textbf{86.21}      & \multicolumn{1}{l|}{\textbf{0.171}}    &90.93      & \multicolumn{1}{l|}{0.424}    &190.20      & \multicolumn{1}{l|}{\textbf{0.171}}    &97.17     \\ \hline
\multicolumn{1}{c|}{\multirow{3}{*}{6}}  & 0.065   & \multicolumn{1}{l|}{0.467}    &170.45      & \multicolumn{1}{l|}{0.642}    &356.78      & \multicolumn{1}{l|}{0.417}    &\textbf{142.16}      & \multicolumn{1}{l|}{0.427}    &152.74      & \multicolumn{1}{l|}{0.653}    &408.34      & \multicolumn{1}{l|}{\textbf{0.403}}    &151.61      \\ 
\multicolumn{1}{l|}{}                    & 0.175   & \multicolumn{1}{l|}{0.273}    &378.87      & \multicolumn{1}{l|}{0.370}    &549.19      & \multicolumn{1}{l|}{\textbf{0.244}}    &328.65      & \multicolumn{1}{l|}{0.254}    &\textbf{328.59}      & \multicolumn{1}{l|}{0.426}    &579.68      & \multicolumn{1}{l|}{0.248}    &356.94      \\
\multicolumn{1}{l|}{}                    & 0.300   & \multicolumn{1}{l|}{0.169}    &809.82      & \multicolumn{1}{l|}{0.245}  &  1003.13     & \multicolumn{1}{l|}{0.159}    &747.84      & \multicolumn{1}{l|}{\textbf{0.153}}    &\textbf{707.79 }     & \multicolumn{1}{l|}{0.266}    &1049.08      & \multicolumn{1}{l|}{0.181}    &797.98      \\ \hline
\multicolumn{1}{c|}{\multirow{3}{*}{13}} & 0.065   & \multicolumn{1}{l|}{0.573}    &620.44      & \multicolumn{1}{l|}{0.675}    &1150.74      & \multicolumn{1}{l|}{0.560}    &549.20      & \multicolumn{1}{l|}{\textbf{0.550}}    &\textbf{527.33 }     & \multicolumn{1}{l|}{0.697}    &1280.02      & \multicolumn{1}{l|}{0.568}    &587.83      \\
\multicolumn{1}{l|}{}                    & 0.175   & \multicolumn{1}{l|}{0.299}    &1461.46      & \multicolumn{1}{l|}{0.372}    &1826.33      & \multicolumn{1}{l|}{0.270}    &1224.43      & \multicolumn{1}{l|}{\textbf{0.260}}    &\textbf{1148.91 }     & \multicolumn{1}{l|}{0.370}    &2030.60      & \multicolumn{1}{l|}{0.271}    &1279.54      \\
\multicolumn{1}{l|}{}                    & 0.300   & \multicolumn{1}{l|}{0.199}    &2744.94      & \multicolumn{1}{l|}{0.241}    &3126.39      & \multicolumn{1}{l|}{0.185}    &2483.97      & \multicolumn{1}{l|}{\textbf{0.179}}    &\textbf{2337.73 }     & \multicolumn{1}{l|}{0.252}    &3349.63     & \multicolumn{1}{l|}{0.192}    &2649.88      \\ \hline
\multicolumn{1}{c|}{\multirow{3}{*}{20}} & 0.065   & \multicolumn{1}{l|}{0.451}    &1236.86      & \multicolumn{1}{l|}{0.530}    &2069.93      & \multicolumn{1}{l|}{\textbf{0.454}}    &1166.22      & \multicolumn{1}{l|}{0.457}    &\textbf{1153.75}      & \multicolumn{1}{l|}{0.581}    &2549.18      & \multicolumn{1}{l|}{0.459}    &1220.78      \\
\multicolumn{1}{l|}{}                    & 0.175   & \multicolumn{1}{l|}{0.268}    &2740.13      & \multicolumn{1}{l|}{0.319}    &3423.48      & \multicolumn{1}{l|}{0.256}    &2475.73      & \multicolumn{1}{l|}{\textbf{0.247}}    &\textbf{2337.61}      & \multicolumn{1}{l|}{0.337}    &3810.23      & \multicolumn{1}{l|}{0.251}    &2616.52      \\
\multicolumn{1}{l|}{}                    & 0.300   & \multicolumn{1}{l|}{0.189}    &5608.73      & \multicolumn{1}{l|}{0.217}    &6090.51    & \multicolumn{1}{l|}{\textbf{0.171}}    &\textbf{4823.11}      & \multicolumn{1}{l|}{0.172}    &4984.82      & \multicolumn{1}{l|}{0.239}    &6871.25      & \multicolumn{1}{l|}{0.176}    &5284.52      \\ \hline

\multicolumn{2}{l|}{\textbf{\textcolor{black}{Mean}}}  
& \multicolumn{1}{l|}{\textcolor{black}{0.302}} 
& \text{\textcolor{black}{1328.60}} 
& \multicolumn{1}{l|}{\textcolor{black}{0.388}} 
& \text{\textcolor{black}{1657.04}} 
& \multicolumn{1}{l|}{\textcolor{black}{\textbf{0.267}}} 
& \text{\textcolor{black}{1173.57}} 
& \multicolumn{1}{l|}{\textcolor{black}{0.268}} 
& \text{\textcolor{black}{1153.23}} 
& \multicolumn{1}{l|}{\textcolor{black}{0.436}} 
& \text{\textcolor{black}{1862.09}} 
& \multicolumn{1}{l|}{\textcolor{black}{\text{0.271}}} 
& \text{\textcolor{black}{1258.72}} \\ \hline

\end{tabular}
\end{adjustbox}
\end{table}

\begin{figure}[!htp]
    \centering
    \begin{subfigure}[b]{0.48\textwidth} 
        \centering
        \includegraphics[width=\textwidth]{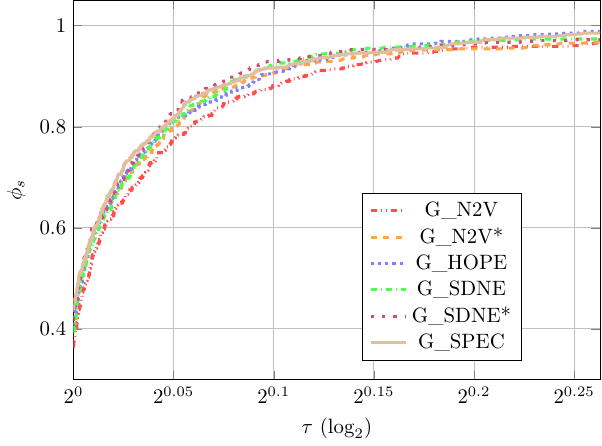}
        \caption{Performance profiles of the \textcolor{black}{gap}.}
        \label{fig:dolangap}
    \end{subfigure}
    \hspace{0.02\textwidth}
    \begin{subfigure}[b]{0.48\textwidth} 
        \centering
        \includegraphics[width=\textwidth]{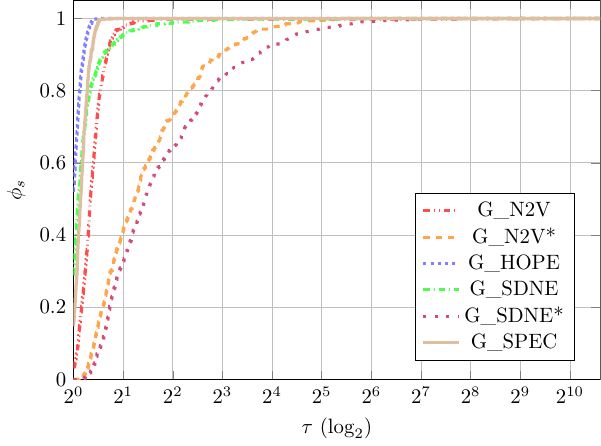}
        \caption{Performance profiles of the \textcolor{black}{Time}.}
        \label{fig:dolantime}
    \end{subfigure}
    \begin{subfigure}[b]{\textwidth}
    \centering
    \includegraphics[scale=0.8]{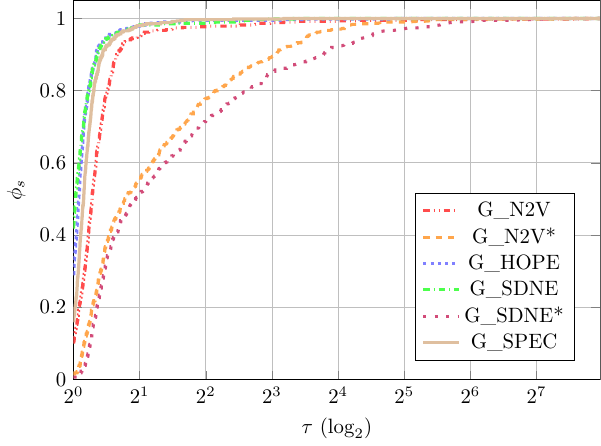}
\caption{Performance profiles of the \textcolor{black}{NPI}.}\label{fig:dolanNPI}
   \end{subfigure}
\caption{Performance profiles \citep{Dolan2002} comparing the \textcolor{black}{gap} and \textcolor{black}{NPI} values per instance.}\label{fig:PP}
 \end{figure}

According to Table~\ref{table:averageresults_npi},  G\_SDNE achieved the best \textcolor{black}{average NPI} in 7 classes of instances, whereas G\_HOPE obtained the best \textcolor{black}{average NPI} in 4 groups. The \textcolor{black}{average NPI} of G\_SPEC was better in only 2 classes of instances, whereas the other methods did not present the best average values. These results demonstrate a better performance of G\_SDNE and G\_HOPE. All methods had an \textcolor{black}{average NPI} greater than 0.4 in the classes instances with the lowest density, i.e. when $\rho=0.065$ and more than 2 layers. {\color{black}Additionally, the mean results indicate that G\_HOPE and G\_SDNE achieved the best and second-best NPI values, respectively.} Figure~\ref{fig:PP} displays the performance profiles of the gap and NPI values.  

One can observe in Figure~\ref{fig:dolangap}, regarding the gap values, despite G\_SDNE* being the dominating curve, those representing G\_HOPE, G\_SDNE and G\_SPEC are substantially close. Considering Figure~\ref{fig:dolantime}, G\_HOPE and G\_SPEC perform better than the other methods, even though G\_SDNE and G\_N2V are highly competitive. In particular, considering $\tau=1$, when we estimate the proportion of cases that a given algorithm achieved the best solution, G\_SPEC was the one that obtained a greater proportion, 0.44, followed by G\_N2V*, G\_SDNE*, G\_HOPE and G\_SDNE, respectively 0.43, 0.42, 0.41 and 0.39. However, when considering the NPI values, by Figure~\ref{fig:dolanNPI}, G\_SDNE, G\_HOPE and G\_SPEC outperformed the other methods, being G\_SDNE* and G\_N2V* significantly worse. In this case, when considering $\tau=1$, G\_SDNE achieved the best results in 0.42 of the instances, whereas G\_HOPE was the best in 0.30.  G\_SPEC and G\_N2V were the best in only 0.16 and 0.10 of the instances, respectively.

 The next section presents the comparative analysis of the two best GL-GRASP heuristics according to the NPI analysis, G\_SDNE and G\_HOPE, and the literature GRASP heuristics.

\subsection{Second Experiment}

This section presents an analysis of G\_SDNE and G\_HOPE contrasting their results with those achieved by the state-of-the-art GRASP heuristics for the C-IGDP, the GRASP2 and GRASP3 algorithms.  We ran each heuristic  30 times for each tested instance and reported the average results.  

The methods are evaluated regarding their \textcolor{black}{gap}, \textcolor{black}{time} and \textcolor{black}{NPI averaged over} groups of instances classified according to different instance features.  Table \ref{table:averageresultscomparative} shows the average gaps and times of the instances grouped according to their number of layers ($\Lambda$) and density values ($\rho$). The values in bold are the best per class of instances.


\begin{table}[!htp]
\caption{Table presenting the \textcolor{black}{average} gaps, times and NPI  by groups of instances with 2, 6, 13 and 20 layers and the tree considered density values.}\label{table:averageresultscomparative}
\begin{adjustbox}{max width=\textwidth}
\begin{tabular}{ll|lll|lll|lll|lll}
\hline
\multicolumn{2}{c|}{Instances} & \multicolumn{3}{c|}{GRASP2} & \multicolumn{3}{c|}{GRASP3} & \multicolumn{3}{c|}{G\_HOPE} & \multicolumn{3}{c}{G\_SDNE} \\ \hline
\multicolumn{1}{l|}{$\Lambda$} & $\rho$ & \multicolumn{1}{l|}{\textcolor{black}{gap}} & \multicolumn{1}{l|}{\textcolor{black}{time}} &\textcolor{black}{NPI} & \multicolumn{1}{l|}{\textcolor{black}{gap}} & \multicolumn{1}{l|}{\textcolor{black}{time}} &\textcolor{black}{NPI} & \multicolumn{1}{l|}{\textcolor{black}{gap}} & \multicolumn{1}{l|}{\textcolor{black}{time}} &\textcolor{black}{NPI} & \multicolumn{1}{l|}{\textcolor{black}{gap}} & \multicolumn{1}{l|}{\textcolor{black}{time}} &\textcolor{black}{NPI} \\ \hline

\multicolumn{1}{c|}{\multirow{3}{*}{2}} & 0.065   
&\multicolumn{1}{l|}{0.241} &\multicolumn{1}{l|}{\textbf{12.53}} &\textbf{0.250} &\multicolumn{1}{l|}{\textbf{0.187}} &\multicolumn{1}{l|}{26.04} &0.327 &\multicolumn{1}{l|}{0.252} &\multicolumn{1}{l|}{14.31} &0.277  &\multicolumn{1}{l|}{0.244} &\multicolumn{1}{l|}{20.17} &0.274 \\ 
\multicolumn{1}{l|}{}& 0.175 
&\multicolumn{1}{l|}{0.228} &\multicolumn{1}{l|}{\textbf{36.91}} &0.321 &\multicolumn{1}{l|}{0.091} &\multicolumn{1}{l|}{93.52} &0.304 &\multicolumn{1}{l|}{\textbf{0.089}} &\multicolumn{1}{l|}{40.95} &0.202  &\multicolumn{1}{l|}{0.175} &\multicolumn{1}{l|}{48.40} &0.289 \\ 
\multicolumn{1}{l|}{} & 0.300   
&\multicolumn{1}{l|}{0.244} &\multicolumn{1}{l|}{\textbf{78.11}} &0.338 &\multicolumn{1}{l|}{\textbf{0.065}} &\multicolumn{1}{l|}{212.16} &0.281 &\multicolumn{1}{l|}{0.183} &\multicolumn{1}{l|}{86.21} &0.279  &\multicolumn{1}{l|}{0.130} &\multicolumn{1}{l|}{90.93} &\textbf{0.232}
\\ \hline

\multicolumn{1}{c|}{\multirow{3}{*}{6}} & 0.065   
&\multicolumn{1}{l|}{0.505} &\multicolumn{1}{l|}{\textbf{133.20}} &0.533 &\multicolumn{1}{l|}{\textbf{0.381}} &\multicolumn{1}{l|}{241.74} &0.499 &\multicolumn{1}{l|}{0.434} &\multicolumn{1}{l|}{142.16} &\textbf{0.469}  &\multicolumn{1}{l|}{0.441} &\multicolumn{1}{l|}{152.74} &0.479 \\ 
\multicolumn{1}{l|}{}& 0.175 
&\multicolumn{1}{l|}{0.326} &\multicolumn{1}{l|}{\textbf{305.91}} &0.408 &\multicolumn{1}{l|}{\textbf{0.095}} &\multicolumn{1}{l|}{612.00} &0.255 &\multicolumn{1}{l|}{0.157} &\multicolumn{1}{l|}{328.65} &\textbf{0.251}  &\multicolumn{1}{l|}{0.170} &\multicolumn{1}{l|}{328.59} &0.260 \\ 
\multicolumn{1}{l|}{} & 0.300   
&\multicolumn{1}{l|}{0.134} &\multicolumn{1}{l|}{\textbf{704.94}} &0.215 &\multicolumn{1}{l|}{\textbf{0.049}} &\multicolumn{1}{l|}{1459.25} & 0.230 &\multicolumn{1}{l|}{0.066} &\multicolumn{1}{l|}{747.84} &0.155  &\multicolumn{1}{l|}{0.063} &\multicolumn{1}{l|}{707.79} &\textbf{0.148}
\\ \hline

\multicolumn{1}{c|}{\multirow{3}{*}{13}} & 0.065   
&\multicolumn{1}{l|}{0.781} &\multicolumn{1}{l|}{\textbf{523.06}} &0.710 &\multicolumn{1}{l|}{\textbf{0.584}} &\multicolumn{1}{l|}{682.88} &\textbf{0.562} &\multicolumn{1}{l|}{0.703} &\multicolumn{1}{l|}{549.20} &0.638  &\multicolumn{1}{l|}{0.698} &\multicolumn{1}{l|}{527.33} &0.625 \\ 
\multicolumn{1}{l|}{}& 0.175 
&\multicolumn{1}{l|}{0.322} &\multicolumn{1}{l|}{\textbf{1145.24}} &0.361 &\multicolumn{1}{l|}{0.294} &\multicolumn{1}{l|}{1704.84} &0.398 &\multicolumn{1}{l|}{0.235} &\multicolumn{1}{l|}{1224.43} &0.286  &\multicolumn{1}{l|}{\textbf{0.234}} &\multicolumn{1}{l|}{1148.91} &\textbf{0.273}\\ 
\multicolumn{1}{l|}{} & 0.300   
&\multicolumn{1}{l|}{0.139} &\multicolumn{1}{l|}{\textbf{2189.97}} &0.203 &\multicolumn{1}{l|}{0.139} &\multicolumn{1}{l|}{3838.05} &0.273 &\multicolumn{1}{l|}{\textbf{0.118}} &\multicolumn{1}{l|}{2483.97} &0.193  &\multicolumn{1}{l|}{\textbf{0.118}} &\multicolumn{1}{l|}{2337.73} &\textbf{0.186}
\\ \hline

\multicolumn{1}{c|}{\multirow{3}{*}{20}} & 0.065   
&\multicolumn{1}{l|}{0.787} &\multicolumn{1}{l|}{\textbf{1042.11}} & 0.554 &\multicolumn{1}{l|}{\textbf{0.698}} &\multicolumn{1}{l|}{1361.85} &0.512 &\multicolumn{1}{l|}{0.728} &\multicolumn{1}{l|}{1166.22} &\textbf{0.506}  &\multicolumn{1}{l|}{0.734} &\multicolumn{1}{l|}{1153.75} &0.511 \\ 
\multicolumn{1}{l|}{}& 0.175 
&\multicolumn{1}{l|}{0.305} &\multicolumn{1}{l|}{2401.97} &0.282 &\multicolumn{1}{l|}{0.336} &\multicolumn{1}{l|}{3376.11} &0.361 &\multicolumn{1}{l|}{0.281} &\multicolumn{1}{l|}{2475.73} &0.264  &\multicolumn{1}{l|}{\textbf{0.276}} &\multicolumn{1}{l|}{\textbf{2337.61}} &\textbf{0.254} \\ 
\multicolumn{1}{l|}{} & 0.300   
&\multicolumn{1}{l|}{0.159} &\multicolumn{1}{l|}{\textbf{4777.90}} &0.217 &\multicolumn{1}{l|}{0.166} &\multicolumn{1}{l|}{7446.10} &0.283 &\multicolumn{1}{l|}{\textbf{0.115}} &\multicolumn{1}{l|}{4823.11} &\textbf{0.177}  &\multicolumn{1}{l|}{0.117} &\multicolumn{1}{l|}{4984.82} &0.178\\ \hline

\multicolumn{2}{l|}{\textcolor{black}{\text{Mean}}} & 
\multicolumn{1}{l|}{\textcolor{black}{\text{0.348}}} & \multicolumn{1}{l|}{\textcolor{black}{\textbf{1112.65}}} & \multicolumn{1}{l|}{\textcolor{black}{\text{0.366}}} & \multicolumn{1}{l|}{\textcolor{black}{\textbf{0.257}}} & \multicolumn{1}{l|}{\textcolor{black}{\text{1754.55}}} & \multicolumn{1}{l|}{\textcolor{black}{\text{0.357}}} & \multicolumn{1}{l|}{\textcolor{black}{\text{0.280}}} & \multicolumn{1}{l|}{\textcolor{black}{\text{1173.57}}} & \multicolumn{1}{l|}{\textcolor{black}{\textbf{0.308}}} & \multicolumn{1}{l|}{\textcolor{black}{\text{0.283}}} & \multicolumn{1}{l|}{\textcolor{black}{\text{1153.23}}} & \multicolumn{1}{l}{\textcolor{black}{\textbf{0.309}}} \\ \hline

\end{tabular}
\end{adjustbox}
\end{table}



On the one hand, one can observe in Table \ref{table:averageresultscomparative} that GRASP3 achieved the lowest \textcolor{black}{average gap} values in most of the instances, encompassing those with 2 and 6 layers and all instances with density values equal to 0.065. On the other hand, in comparison to GRASP3, the GL-GRASP heuristics achieved better \textcolor{black}{average gap} values for the instances with 13 and 20 layers (except when the density was 0.065) in inferior computational times. In comparison to the other methods, the \textcolor{black}{average gap} values obtained by GRASP2 were worse despite its computational times being better in most of the cases. 

Table \ref{table3_d1} shows a summary of the average results of the \textcolor{black}{gap} and \textcolor{black}{NPI} per group of instances with the same density ($\rho \in\{0.065, 0.175, 0.300\}$) and number of layers ($\Lambda\in \{2, 6, 13, 20\}$). It presents the classification of the methods according to these metrics. 

\begin{table}[!htp]
\centering
\caption{Ranking of the heuristics according to their \textcolor{black}{average gap} and \textcolor{black}{NPI}.}
\begin{adjustbox}{max width=1.0\textwidth}
\begin{tabular}{ c|c|c|c|c|c|c|c } 
\hline
\multirow{3}{*}{$\Lambda$}&\multirow{3}{*}{Class.}&\multicolumn{6}{c}{$\rho$} \\ \cline{3-8}
 &  & \multicolumn{2}{c|}{0.065} & \multicolumn{2}{c|}{0.175} & \multicolumn{2}{c}{0.300} \\ \cline{3-8}
 &  & \textcolor{black}{gap} & \textcolor{black}{NPI} & \textcolor{black}{gap} & \textcolor{black}{NPI} & \textcolor{black}{gap} & \textcolor{black}{NPI}\\ \hline
\multirow{3}{*}{2} 
 & \nth{1} &GRASP3 (0.187)   &GRASP2 (0.250)  &\cellcolor{lgray} G\_HOPE (0.089) &\cellcolor{lgray} G\_HOPE (0.202) &GRASP3 (0.065) &\cellcolor{lightgray} G\_SDNE (0.232)\\
 & \nth{2} &GRASP2 (0.241)   &\cellcolor{lightgray} G\_SDNE (0.274)  &GRASP3 (0.091) &\cellcolor{lightgray} G\_SDNE (0.289) &\cellcolor{lightgray} G\_SDNE (0.130) &\cellcolor{lgray} G\_HOPE (0.279)\\
 & \nth{3} &\cellcolor{lightgray} G\_SDNE (0.244)  &\cellcolor{lgray} G\_HOPE (0.277)  &\cellcolor{lightgray} G\_SDNE (0.175) &GRASP3 (0.304) &\cellcolor{lgray} G\_HOPE (0.183) &GRASP3 (0.281)\\
 & \nth{4} &\cellcolor{lgray} G\_HOPE (0.252)  &GRASP3 (0.327)  &GRASP2 (0.228)     &GRASP2 (0.321) &GRASP2 (0.244) &GRASP2 (0.338)\\ \hline

\multirow{3}{*}{6} 
 & \nth{1} &GRASP3 (0.381)                          &\cellcolor{lgray} G\_HOPE (0.469)  &GRASP3 (0.095)                         &\cellcolor{lgray} G\_HOPE (0.251) &GRASP3 (0.049) &\cellcolor{lightgray} G\_SDNE (0.148)\\
 & \nth{2} &\cellcolor{lgray} G\_HOPE (0.434)      &\cellcolor{lightgray} G\_SDNE (0.479)  &\cellcolor{lgray} G\_HOPE (0.157)     &GRASP3 (0.255) &\cellcolor{lightgray} G\_SDNE (0.063) &\cellcolor{lgray} G\_HOPE (0.155)\\
 & \nth{3} &\cellcolor{lightgray} G\_SDNE (0.441)  &GRASP3 (0.499)  &\cellcolor{lightgray} G\_SDNE (0.170) &\cellcolor{lightgray} G\_SDNE (0.260) &\cellcolor{lgray} G\_HOPE (0.066) &GRASP2 (0.215)\\
 & \nth{4} &GRASP2 (505)                            &GRASP2 (0.533)  &GRASP2 (0.326)                         &GRASP2 (0.408) &GRASP2 (0.134) &GRASP3 (0.230)\\ \hline

 \multirow{3}{*}{13} 
 & \nth{1} &GRASP3 (0.584)                          &GRASP3 (0.562)  &\cellcolor{lightgray} G\_SDNE (0.234) &\cellcolor{lightgray} G\_SDNE (0.273) &\cellcolor{lgray} G\_HOPE (0.1176) &\cellcolor{lightgray} G\_SDNE (0.186)\\
 & \nth{2} &\cellcolor{lightgray} G\_SDNE (0.698)  &\cellcolor{lightgray} G\_SDNE (0.625)  &\cellcolor{lgray} G\_HOPE (0.235)     &\cellcolor{lgray} G\_HOPE (0.286) &\cellcolor{lightgray} G\_SDNE (0.1182) &\cellcolor{lgray} G\_HOPE (0.193)\\
 & \nth{3} &\cellcolor{lgray} G\_HOPE (0.703)      &\cellcolor{lgray} G\_HOPE (0.638)  &GRASP3 (0.294)                         &GRASP2 (0.631) &GRASP3 (0.1389) &GRASP2 (0.203)\\
 & \nth{4} &GRASP2 (0.781)                          &GRASP2 (0.710)  &GRASP2 (0.322)                         &GRASP3 (0.398) &GRASP2 (0.1391) &GRASP3 (0.273)\\ \hline

 \multirow{3}{*}{20} 
 & \nth{1} &GRASP3 (0.698)                          &\cellcolor{lgray} G\_HOPE (0.506)  &\cellcolor{lightgray} G\_SDNE (0.276)  &\cellcolor{lightgray} G\_SDNE (0.254) &\cellcolor{lgray} G\_HOPE (0.115) &\cellcolor{lgray} G\_HOPE (0.177)\\
 & \nth{2} &\cellcolor{lgray} G\_HOPE (0.728)      &\cellcolor{lightgray} G\_SDNE (0.511)  &\cellcolor{lgray} G\_HOPE (0.281)      &\cellcolor{lgray} G\_HOPE (0.264) &\cellcolor{lightgray} G\_SDNE (0.117) &\cellcolor{lightgray} G\_SDNE (0.178)\\
 & \nth{3} &\cellcolor{lightgray} G\_SDNE (0.734)  &GRASP3 (0.512)  &GRASP2 (0.305)                          &GRASP2 (0.282) &GRASP2 (0.159) &GRASP2 (0.217)\\
 & \nth{4} &GRASP2 (0.787)                          &GRASP2 (0.554)  &GRASP3 (0.336)                          &GRASP3 (0.361) &GRASP3 (0.166) &GRASP3 (0.283)\\ \hline

\end{tabular}
\end{adjustbox}
\label{table3_d1}
\end{table}

The results indicate that G\_HOPE and G\_SDNE achieved the best or second best \textcolor{black}{gap} and \textcolor{black}{NPI} in all case tests for $\rho$ equals 0.175 and 0.300, respectively. Considering \textcolor{black}{NPI}, G\_SDNE and G\_HOPE had the best or second best results in all groups of instances whose $\rho$ equals $0.065$ and 0.300, respectively. Still, G\_SDNE achieved the best \textcolor{black}{gap} and \textcolor{black}{NPI} in all the cases with $\rho=0.175$ and 13 and 20 layers, for which G\_HOPE ranked the second best. In summary, these results indicate that the GL-GRASP heuristics were better in comparison to the literature GRASP heuristics on instances with more layers and higher density. Moreover, they suggest that for the other instances, these algorithms were highly competitive.



\begin{figure}[!htp]
  \begin{subfigure}{.5\linewidth}
    \centering   
    \includegraphics[width=\linewidth]{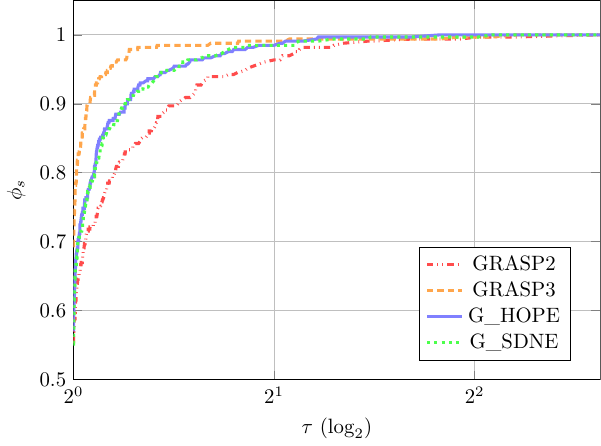}
    \caption{Performance profiles of instances with $\lambda=2\wedge6$, displaying their \textcolor{black}{gap} per instance.}

  \end{subfigure}%
  \begin{subfigure}{.5\linewidth}
    \centering
    \includegraphics[width=\linewidth]{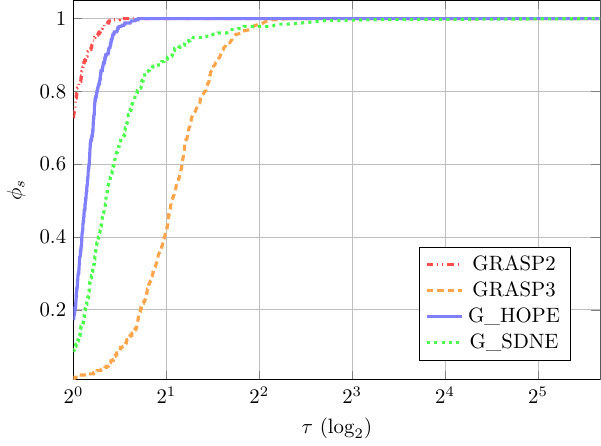}
    \caption{Performance profiles of instances with $\lambda=2\wedge6$, displaying their \textcolor{black}{time} per instance.}
  \end{subfigure}%

  \medskip

  \begin{subfigure}{.5\linewidth}
    \centering
    \includegraphics[width=\linewidth]{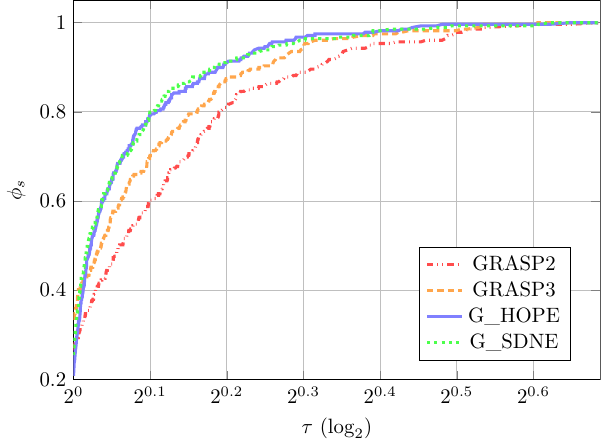}
    \caption{Performance profiles of instances with $\lambda=13\wedge20$, displaying their \textcolor{black}{gap} per instance.}
  \end{subfigure}%
  \begin{subfigure}{.5\linewidth}
    \centering
    \includegraphics[width=\linewidth]{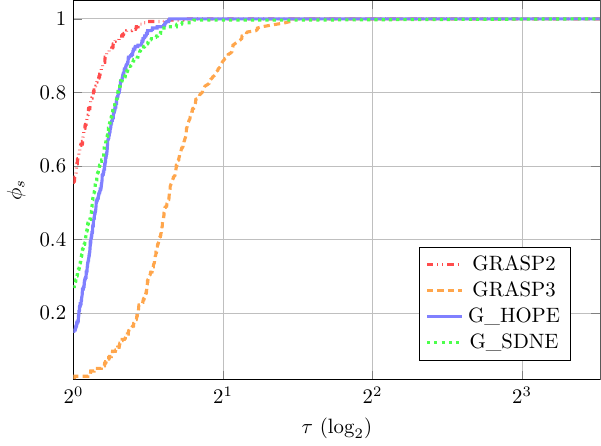}
    \caption{Performance profiles of instances with $\lambda=13\wedge20$, displaying their \textcolor{black}{time} per instance.}
  \end{subfigure}%

  \caption{Performance profiles of all methods on instances separated by the number of layers.}\label{fig_perfgaptimr}
\end{figure}

By observing the better performance of the GL-GRASP heuristics on instances with more layers, we further investigate these instances by plotting the performance profiles of the instances according to their number of layers.  Figure~\ref{fig_perfgaptimr} shows the performance profiles of the gap and time of the four algorithms divided into different groups of instances by the number of layers. 

Figure~\ref{fig_perfgaptimr} shows a better performance of GRASP3 on the \textcolor{black}{gap} in instances where $\Lambda=2\wedge 6$. According to these results for the graphs with the highest number of layers, $\Lambda=13\wedge 20$, the GL-GRASP heuristics outperformed the other methods considering the \textcolor{black}{gap} values. By analyzing the performance profiles of the times, the GL-GRASP heuristics outperform GRASP3 in both groups of instances. Despite GRASP2 achieving the worst \textcolor{black}{gap} performance for all instance groups, its time performance profile outperformed the other heuristics in both groups of instances. 

\begin{figure}[!htp]
  \begin{subfigure}{.5\linewidth}
    \centering   
    \includegraphics[width=\linewidth]{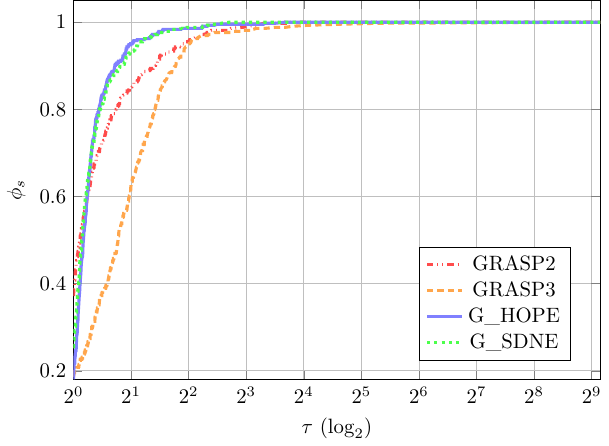}
    \caption{Performance profiles of instances with $\Lambda=2\wedge6$, displaying their \textcolor{black}{NPI} per instance.}\label{fig_perf2a}

  \end{subfigure}%
  \begin{subfigure}{.5\linewidth}
    \centering
    \includegraphics[width=\linewidth]{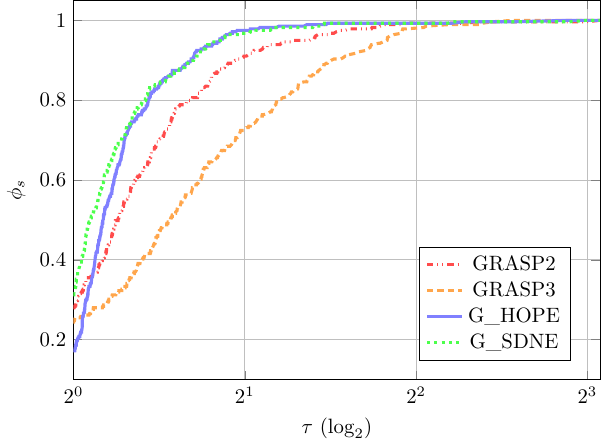}
    \caption{Performance profiles of instances with $\Lambda=13\wedge20$, displaying their \textcolor{black}{NPI} per instance.}\label{fig_perf2b}
  \end{subfigure}%

  \caption{Performance profiles \citep{Dolan2002} comparing the NPI values of the instances separated according to the number of layers.}\label{fig_perf2}
\end{figure}

For a better notion of the proportion of instances for which the GRASP heuristics performed better considering the \textcolor{black}{NPI} values, Figure~\ref{fig_perf2} displays the performance profiles of the NPI of the methods under evaluation.


It is possible to observe that G\_HOPE and G\_SDNE had similar behavior and that GRASP2 and GRASP3 were dominated by the GL-GRASP curves. On the one hand, according to Figure~\ref{fig_perf2a}, for $\Lambda = 2\wedge6$, GRASP2 achieved the best results when $\tau = 1$, meaning that it had the best NPI in 0.375 of the instances against the 0.255, 0.189, 0.182 of G\_SDNE,  GRASP3 and G\_HOPE, respectively. However, when  $\tau = 1.173$, both GL-GRASP heuristics performed better than GRASP2. On the other hand, Figure~\ref{fig_perf2b} demonstrates that G\_SDNE outperformed the other methods within the whole interval. It achieved the best results in 0.312 of the instances, against 0.280, 0.240 and 0.168 obtained by GRASP2, GRASP3 and  G\_HOPE, respectively. G\_HOPE surpassed   GRASP2 and GRASP3 when $\tau= $1.083, obtaining  NPI values competitive with G\_SDNE.

 Table \ref{tableopt2} displays the ranking of the methods according to the percentage of optimal solutions per group of instances divided according to their density and number of layers. According to this table, GRASP3 obtained a higher percentage of optimal solutions for instances with 2 and 6 layers and in all instances where $\rho$ is $0.06$ and $0.17$. However, G\_SDNE and G\_HOPE found more optimal solutions in the graphs with the highest density and the two largest number of layers. The Supplementary Material presents details on the optimal solution values for each tested instance.
\begin{table}[!htp]
\centering
\caption{Percentage of optimal solutions achieved by the GRASP heuristics.}\label{tableopt2}
\begin{adjustbox}{max width=0.8\textwidth}
\begin{tabular}{ c|c|c|c|c } \hline
\multicolumn{5}{ l }{Percentage of Optimal Solutions} \\ \hline

\multirow{2}{*}{$\Lambda$}&\multirow{2}{*}{Class.}&\multicolumn{3}{c}{$\rho$} \\ \cline{3-5}
 &  & \multicolumn{1}{c|}{0.065} & \multicolumn{1}{c|}{0.175} & \multicolumn{1}{c}{0.300} \\ \hline
 \multirow{3}{*}{2} 
 & \nth{1} &GRASP3 (94.74)  &GRASP3 (100.00) &GRASP3 (98.25) \\
 & \nth{2} &\cellcolor{lightgray} G\_SDNE (92.98)  &\cellcolor{lgray} G\_HOPE (96.49) &\cellcolor{lightgray} G\_SDNE (96.49) \\
 & \nth{3} &\cellcolor{lgray} G\_HOPE (91.23)  &\cellcolor{lightgray} G\_SDNE (94.74) &\cellcolor{lgray} G\_HOPE (91.23) \\
 & \nth{4} &GRASP2 (89.47)  &GRASP2 (92.98) &GRASP2 (89.47) \\ \hline 

\multirow{4}{*}{6} 
 & \nth{1} &GRASP3 (58.49)  &GRASP3 (79.25) &GRASP3 (84.91) \\
 & \nth{2} &\cellcolor{lgray} G\_HOPE, \cellcolor{lightgray} G\_SDNE (54.72)   &\cellcolor{lightgray} G\_SDNE (69.81) &\cellcolor{lightgray} G\_SDNE (77.36) \\
 & \nth{3} &GRASP2 (49.06)  &\cellcolor{lgray} G\_HOPE (67.92) &\cellcolor{lgray} G\_HOPE (75.47) \\
 & \nth{4} &---  &GRASP2 (43.40) &GRASP2 (58.49) \\ \hline

\multirow{4}{*}{13} 
 & \nth{1} &GRASP3 (31.91)  &GRASP3 (34.04) &GRASP3, \cellcolor{lightgray} G\_SDNE (31.91)  \\
 & \nth{2} &\cellcolor{lgray} G\_HOPE, \cellcolor{lightgray} G\_SDNE(29.79)   &\cellcolor{lgray} G\_HOPE (29.79) &\cellcolor{lgray} G\_HOPE (27.66) \\
 & \nth{3} &GRASP2 (25.53)  &\cellcolor{lightgray} G\_SDNE (27.66) &GRASP2 (21.28) \\
 & \nth{4} &---  &GRASP2 (23.40) &--- \\ \hline

 \multirow{5}{*}{20} 
 & \nth{1} &GRASP3 (26.09)  &GRASP3 (32.61) &\cellcolor{lightgray} G\_SDNE(23.91) \\
 & \nth{2} &GRASP2, \cellcolor{lgray} G\_HOPE, \cellcolor{lightgray} G\_SDNE (19.57)    &\cellcolor{lgray} G\_HOPE (17.39) &\cellcolor{lgray} G\_HOPE (19.57) \\
 & \nth{3} &---  &\cellcolor{lightgray} G\_SDNE(15.22) &GRASP3 (17.39) \\
 & \nth{4} &---  &GRASP2 (10.87) &GRASP2 (15.22) \\ \hline

\end{tabular}
\end{adjustbox}

\end{table}



\textcolor{black}{
Finally, statistical significance is evaluated for both the solution gap and runtime using pairwise Wilcoxon signed-rank tests at the 5\% significance level. Table~\ref{tab:wilcoxon} presents the results; a dagger symbol ($\dagger$) denotes statistically significant differences between the compared methods.}


\begin{table}[ht]
\centering
\color{black}
\caption{\textcolor{black}{Wilcoxon signed-rank test p-values: upper triangle shows gap values; lower triangle shows runtime values. A $\dagger$ indicates a statistically significant difference between methods ($p < 0.05$).}}
\label{tab:wilcoxon}
\begin{tabular}{lcccc}
\toprule
        & GRASP2 & GRASP3 & G\_HOPE & G\_SDNE \\
\midrule
\rowcolor{white} 
GRASP2 & -- 
        & \cellcolor{lightergray}$6.07 \times 10^{-28^{\dagger}}$ & \cellcolor{lightergray}$1.99 \times 10^{-23^{\dagger}}$ & \cellcolor{lightergray}$3.80 \times 10^{-24^{\dagger}}$ \\
        
\rowcolor{white} 
GRASP3 & \cellcolor{lightgray}$3.50 \times 10^{-95^{\dagger}}$ & -- 
        & \cellcolor{lightergray}$6.19 \times 10^{-1}$ 
        & \cellcolor{lightergray}$7.00 \times 10^{-1}$ \\
        
\rowcolor{white} 
G\_HOPE & \cellcolor{lightgray}$3.08 \times 10^{-28^{\dagger}}$ 
& \cellcolor{lightgray}$3.54 \times 10^{-94^{\dagger}}$ & -- 
        & \cellcolor{lightergray}$9.47 \times 10^{-1}$ \\
        
\rowcolor{white} 
G\_SDNE & \cellcolor{lightgray}$1.64 \times 10^{-20^{\dagger}}$ 
& \cellcolor{lightgray}$6.40 \times 10^{-87^{\dagger}}$ 
& \cellcolor{lightgray}$3.15 \times 10^{-1}$ 
        & -- \\
\bottomrule
\end{tabular}
\end{table}

{\color{black}

{\color{black}The pairwise comparison of the methods highlights significant differences in both terms of gap and runtime. First, the gap values obtained by GRASP2 are statistically different from those of GRASP3, G\_HOPE, and G\_SDNE. However, no significant differences in gap were found when comparing GRASP3, G\_HOPE, and G\_SDNE, indicating that these three methods achieved similar solution quality.

Regarding runtime, GRASP2 is significantly faster than the other methods. In contrast, GRASP3 is notably slower than GRASP2, G\_HOPE, and G\_SDNE. Furthermore, G\_HOPE and G\_SDNE did not exhibit significant differences in their computational times.}

Table~\ref{table_c_ph} provides an overall summary of the solution quality, including the average objective solution values ($\overline{C}$) and gaps of the solutions obtained by the construction phases of all GRASP heuristics evaluated in this section. $C_{\text{HOPE}}$ and $C_{\text{SDNE}}$ refer to the construction phases of G\_HOPE and G\_SDNE, respectively. This table also shows the  $\overline{C}$ and gaps of the solutions achieved by the GRASP heuristics.}

\begin{table}[!htp]
\centering
\color{black}
\caption{\textcolor{black}{Average crossings ($\overline{C}$), gaps, and execution times (in seconds) for the constructive and final phases.}}
\label{table_c_ph}
\begin{adjustbox}{max width=\textwidth}
\begin{tabular}{c | c | c || c | c | c | c} 
\hline

Method & $\overline{C}$ & gap  & Method & $\overline{C}$ & gap & time (s) \\ \hline
$C_2$  &18614.68  &12.876  & GRASP2 &17199.17  &0.341 &1030.30   \\
$C_3$  &18022.55   &5.846   & GRASP3 &17196.27  &0.247 &1633.13   \\
$C_{\text{HOPE}}$ &18752.13  &15.302  & G\_HOPE &17191.26  &0.273 &1087.28    \\
$C_{\text{SDNE}}$ &18735.68  &15.371  & G\_SDNE &17191.42  &0.276 &1068.43   \\
\hline
\end{tabular}
\end{adjustbox}
\end{table}

{\color{black}As observed in Table \ref{table_c_ph}, $C_3$ performs well in achieving good initial solutions, presenting the lowest average objective function value and gap among all construction phases. However,  the $\overline{C}$ of the solutions obtained by G\_{HOPE} and G\_{SDNE} were lower and required significantly less time than GRASP3, mostly due to its costly $C_3$. The greater time demand observed in $C_3$ stems directly from its more time-consuming approach. At each iteration, the algorithm exhaustively evaluates the insertion cost ($\mathop{\varrho(\nu,p)}$) of each candidate incremental node in all possible positions within its respective layer. This comprehensive process, while intensively searching for higher-quality initial solutions, results in significantly greater computational overhead.

The following section presents the experiment in large-scale instances.
}

\subsection{\textcolor{black}{Third Experiment}}

{\color{black} This section presents the experiment in newly generated instances to evaluate the scalability of the methods.

For all methods, including Gurobi, we imposed a time limit of 10800 seconds per run. However, the GRASP heuristics were allowed to exceed this time limit to complete an ongoing iteration. The results reported in Table~\ref{tab:detailed_results_additional} refer to the average of 10 independent executions. The GRASP parameters $\eta$ and $\eta_{max}$ were fixed at $100$ and $20$, respectively. As Gurobi did not reach optimality within the time limit, the reported gap is relative to the best-known solution (BKS). }
\begin{table}[!htp]
{\color{black}{\footnotesize\caption{{\color{black}Mean gaps and runtimes (in seconds) for GRASP2, GRASP3, G\_HOPE, G\_SDNE, and Gurobi in the newly generated instances.}}\label{tab:detailed_results_additional}
\begin{tabular}{c|c|c c|c c|c c|c c|c c} 
\hline
\multirow{2}{*}{Instance} & 
\multirow{2}{*}{\textbf{$d$}} & 
\multicolumn{2}{c|}{GRASP2} & 
\multicolumn{2}{c|}{GRASP3} & 
\multicolumn{2}{c|}{G\_HOPE} & 
\multicolumn{2}{c|}{G\_SDNE} & 
\multicolumn{2}{c}{Gurobi} \\
\cline{3-12}
& & gap & time & gap & time & gap & time & gap & time & gap & time \\
\hline
\scriptsize incgraph\_2 & \multirow{4}{*}{\textbf{\scriptsize 1}} & 0.001 & 4375.27 & 0.276 & 9012.53 & \textbf{0.000} & \textbf{4100.10} & \textbf{0.000} & 4308.10 & 0.384 & 10836.10 \\
\scriptsize incgraph\_3 & & 0.002 & 10534.60 & 0.855 & 19927.36 & \textbf{0.000} & 6744.92 & \textbf{0.000} & \textbf{5842.59} & 0.751 & 10838.03 \\
\scriptsize incgraph\_4 & & 0.028 & 10833.22 & 0.983 & 17403.80 & 0.014 & 7994.38 & \textbf{0.010} & \textbf{7393.75} & 0.665 & 10817.54 \\
\scriptsize incgraph\_5 & & 0.033 & 7572.01 & 1.384 & 17176.61 & 0.006 & 6835.02 & \textbf{0.005} & \textbf{6633.38} & 0.860 & 10820.69 \\
\hline
\scriptsize incgraph\_2 & \multirow{4}{*}{\textbf{\scriptsize 2}} & 0.003 & 4414.44 & 0.407 & \textbf{4357.18} & \textbf{0.000} & 4710.33 & \textbf{0.000} & 4669.57 & 0.653 & 11233.02 \\
\scriptsize incgraph\_3 & & 0.002 & 8172.45 & 1.180 & 10730.83 & \textbf{0.000} & 8172.01 & \textbf{0.000} & \textbf{7553.50} & --- & --- \\
\scriptsize incgraph\_4 & & 0.005 & 11722.44 & 1.925 & 14245.40 & 0.007 & 11024.17 & \textbf{0.001} & 11081.10 & 0.452 & \textbf{10826.43} \\
\scriptsize incgraph\_5 & & 0.094 & 12827.46 & 2.075 & 18640.18 & \textbf{0.007} & 11732.72 & 0.013 & \textbf{10825.24} & --- & --- \\
\hline
\scriptsize incgraph\_2 & \multirow{4}{*}{\textbf{\scriptsize 3}} & 0.002 & 4632.46 & 0.884 & 5438.06 & \textbf{0.000} & \textbf{4553.38} & \textbf{0.000} & 4613.14 & 0.655 & 11214.32 \\
\scriptsize incgraph\_3 & & 0.004 & 8774.01 & 1.904 & 17639.96 & \textbf{0.001} & \textbf{8251.19} & 0.002 & 8317.33 & --- & --- \\
\scriptsize incgraph\_4 & & \textbf{0.001} & 12772.52 & 2.525 & 29157.01 & 0.012 & 11831.89 & 0.002 & 11758.32 & 0.901 & \textbf{10844.90}\\
\scriptsize incgraph\_5 & & 0.076 & 19037.75 & 3.143 & 16907.58 & \textbf{0.037} & 14503.63 & 0.047 & \textbf{13895.19} & --- & --- \\
\midrule
\textbf{Mean} & & 0.021 & 9639.05 & 1.462 & 15053.04 & \textbf{0.007} & 8371.14 & \textbf{0.007} & 8074.27 & 0.665 & 10928.88 \\
\bottomrule
\end{tabular}}}
\end{table}

According to Table~\ref{tab:detailed_results_additional}, the GL-GRASP heuristics, G\_HOPE and G\_SDNE, outperformed the other methods, presenting an average gap of 0.07. Each of them achieved the best gaps in 8 of the 12 instances. Among the literature heuristics, GRASP2 achieved better results, reaching an average gap of 0.021, achieving the best gap in 1 of the 12 instances. GRASP3 had an average gap of 1.462, performing significantly worse than the other methods.

Regarding computational efficiency, both G\_HOPE (average time: 8371.14s) and G\_SDNE (8074.27s) were considerably faster, on average, than the literature heuristics GRASP2 (9639.05s) and GRASP3 (15053.04s). These results demonstrate the strong efficiency of the learning-based construction phases in G\_HOPE and G\_SDNE for large instances. Therefore, the balance achieved by G\_HOPE and G\_SDNE between solution quality and computation time, observed in the benchmark instances, was more accentuated in more complex large-scale instances.


\section{Conclusions}
\label{sec:sample5}

This paper introduced an innovative incorporation of deep learning strategies in metaheuristics.

Graph Representation Learning (GRL), also known as Representation Learning on Graphs, is a rapidly evolving field with immense potential to transform how we understand and leverage complex data structures. By harnessing the power of graph-based representations, we can uncover hidden patterns and relationships in data that are often invisible to traditional methods. This innovative approach enables us to solve different real-world problems, from predicting molecular interactions for drug discovery to optimizing transportation networks and enhancing social network analysis.

In this paper, we investigated the graph drawing problem known as Constrained Incremental Graph Drawing (C-IGDP), for which GRASP heuristics are state-of-the-art methods. By extracting latent information from the graphs through {four} cutting-edge GRL algorithms, we employ the node embedding information to guide the search process of the proposed metaheuristics. We implemented the state-of-the-art GRASP heuristics as reported in the literature for the C-IGDP and adapted them to incorporate information from the GRL algorithms. The time to extract the node embeddings was considered in the CPU time evaluation for a fair comparison.

In the comparative analysis of the GL\_GRASP versions with different embedding strategies, we found that  G\_HOPE and G\_SDNE heuristics offered the best trade-off between solution quality and computational time. Consequently, these two heuristics were selected for the experiments in which we compare our approach with the literature heuristics, namely GRASP2 and GRASP3.

On the benchmark set, the solution quality of G\_HOPE and G\_SDNE is not statistically different from that of GRASP3, while their runtimes are significantly shorter. Compared with GRASP2, both G\_HOPE and G\_SDNE deliver better solution quality, although GRASP2 attains the shortest runtimes on this set.
On the newly generated large-scale, denser instances, G\_HOPE and G\_SDNE achieve lower average gaps and faster average runtimes than the literature heuristics (GRASP2 and GRASP3). These statements are supported by the Wilcoxon tests.

The results of this paper indicate a promising new approach to solving the C-IGDP. This approach offers valuable insights for addressing similar problems in the literature besides contributing to advancing knowledge in this area. Furthermore, it underscores the importance of integrating emerging machine learning techniques with metaheuristics to help the search for solutions for various combinatorial optimization problems.


\section*{Acknowledgements}
The authors thank the Coordenação de Aperfeiçoamento de Pessoal de Nível Superior/Brasil (CAPES) – Finance Code 001 – for the financial support. The authors are also grateful for the financial support provided by CNPq (403735/2021-1; 309385/2021-0) and FAPESP (2013/07375-0; 2022/05803-3).

Research carried out using the computational resources of the Center for Mathematical Sciences Applied to Industry (CeMEAI) funded by FAPESP (grant 2013/07375-0).







\textcolor{red}{\\ *This is the author’s accepted manuscript. The final version will appear in European Journal of Operational Research. Supplementary material will be available at the journal’s website or upon request.}


\begin{thebibliography}{57}
\expandafter\ifx\csname natexlab\endcsname\relax\def\natexlab#1{#1}\fi
\providecommand{\url}[1]{\texttt{#1}}
\providecommand{\href}[2]{#2}
\providecommand{\path}[1]{#1}
\providecommand{\DOIprefix}{doi:}
\providecommand{\ArXivprefix}{arXiv:}
\providecommand{\URLprefix}{URL: }
\providecommand{\Pubmedprefix}{pmid:}
\providecommand{\doi}[1]{\href{http://dx.doi.org/#1}{\path{#1}}}
\providecommand{\Pubmed}[1]{\href{pmid:#1}{\path{#1}}}
\providecommand{\bibinfo}[2]{#2}
\ifx\xfnm\relax \def\xfnm[#1]{\unskip,\space#1}\fi
\bibitem[{Bachmaier et~al.(2010)Bachmaier, Brandenburg, Brunner and H{\"u}bner}]{Bachmaier2010}
\bibinfo{author}{Bachmaier, C.}, \bibinfo{author}{Brandenburg, F.J.}, \bibinfo{author}{Brunner, W.}, \bibinfo{author}{H{\"u}bner, F.}, \bibinfo{year}{2010}.
\newblock \bibinfo{title}{A global k-level crossing reduction algorithm}, in: \bibinfo{editor}{Rahman, M.S.}, \bibinfo{editor}{Fujita, S.} (Eds.), \bibinfo{booktitle}{WALCOM: Algorithms and Computation}, \bibinfo{publisher}{Springer Berlin Heidelberg}, \bibinfo{address}{Berlin, Heidelberg}. pp. \bibinfo{pages}{70--81}.
\bibitem[{Battista et~al.(1998)Battista, Eades, Tamassia and Tollis}]{batista3.3}
\bibinfo{author}{Battista, G.}, \bibinfo{author}{Eades, P.}, \bibinfo{author}{Tamassia, R.}, \bibinfo{author}{Tollis, I.}, \bibinfo{year}{1998}.
\newblock \bibinfo{title}{Graph Drawing: Algorithms for the Visualization of Graphs}.
\newblock \bibinfo{publisher}{Prentice Hall PTR}, \bibinfo{address}{USA}.
\bibitem[{Bengio et~al.(2021)Bengio, Lodi and Prouvost}]{bengio2021machine}
\bibinfo{author}{Bengio, Y.}, \bibinfo{author}{Lodi, A.}, \bibinfo{author}{Prouvost, A.}, \bibinfo{year}{2021}.
\newblock \bibinfo{title}{Machine learning for combinatorial optimization: a methodological tour d’horizon}.
\newblock \bibinfo{journal}{European Journal of Operational Research} \bibinfo{volume}{290}, \bibinfo{pages}{405--421}.
\bibitem[{Berthold(2013)}]{Timo}
\bibinfo{author}{Berthold, T.}, \bibinfo{year}{2013}.
\newblock \bibinfo{title}{Measuring the impact of primal heuristics}.
\newblock \bibinfo{journal}{Operations Research Letters} \bibinfo{volume}{41}, \bibinfo{pages}{611–614}.
\newblock \DOIprefix\doi{10.1016/j.orl.2013.08.007}.
\bibitem[{Binucci et~al.(2019)Binucci, Brandes, Dwyer, Gronemann, von Hanxleden, van Kreveld, Mutzel, Schaefer, Schreiber and Speckmann}]{Binucci:3.1}
\bibinfo{author}{Binucci, C.}, \bibinfo{author}{Brandes, U.}, \bibinfo{author}{Dwyer, T.}, \bibinfo{author}{Gronemann, M.}, \bibinfo{author}{von Hanxleden, R.}, \bibinfo{author}{van Kreveld, M.}, \bibinfo{author}{Mutzel, P.}, \bibinfo{author}{Schaefer, M.}, \bibinfo{author}{Schreiber, F.}, \bibinfo{author}{Speckmann, B.}, \bibinfo{year}{2019}.
\newblock \bibinfo{title}{10 reasons to get interested in graph drawing}, in: \bibinfo{booktitle}{Computing and Software Science: State of the Art and Perspectives}, \bibinfo{publisher}{Springer International Publishing}, \bibinfo{address}{Cham}. pp. \bibinfo{pages}{85--104}.
\newblock \URLprefix \url{https://doi.org/10.1007/978-3-319-91908-96}.
\bibitem[{Cai et~al.(2018)Cai, Zheng and Chang}]{cai:03}
\bibinfo{author}{Cai, H.}, \bibinfo{author}{Zheng, V.}, \bibinfo{author}{Chang, K.}, \bibinfo{year}{2018}.
\newblock \bibinfo{title}{A comprehensive survey of graph embedding: Problems, techniques, and applications}.
\newblock \bibinfo{journal}{IEEE Transactions on Knowledge and Data Engineering} \bibinfo{volume}{30}, \bibinfo{pages}{1616--1637}.
\bibitem[{Cen et~al.(2023)Cen, Hou, Wang, Chen, Luo, Yu, Zhang, Yao, Zeng, Guo, Dong, Yang, Zhang, Dai, Wang, Zhou, Yang and Tang}]{cen2021cogdl}
\bibinfo{author}{Cen, Y.}, \bibinfo{author}{Hou, Z.}, \bibinfo{author}{Wang, Y.}, \bibinfo{author}{Chen, Q.}, \bibinfo{author}{Luo, Y.}, \bibinfo{author}{Yu, Z.}, \bibinfo{author}{Zhang, H.}, \bibinfo{author}{Yao, X.}, \bibinfo{author}{Zeng, A.}, \bibinfo{author}{Guo, S.}, \bibinfo{author}{Dong, Y.}, \bibinfo{author}{Yang, Y.}, \bibinfo{author}{Zhang, P.}, \bibinfo{author}{Dai, G.}, \bibinfo{author}{Wang, Y.}, \bibinfo{author}{Zhou, C.}, \bibinfo{author}{Yang, H.}, \bibinfo{author}{Tang, J.}, \bibinfo{year}{2023}.
\newblock \bibinfo{title}{Cogdl: A comprehensive library for graph deep learning}, in: \bibinfo{booktitle}{Proceedings of the ACM Web Conference 2023}, \bibinfo{publisher}{Association for Computing Machinery}, \bibinfo{address}{New York, NY, USA}. p. \bibinfo{pages}{747–758}.
\newblock \URLprefix \url{https://docs.cogdl.ai/en/latest/}, \DOIprefix\doi{10.1145/3543507.3583472}.
\bibitem[{Chen et~al.(2022)Chen, Jiang, Yu, Zhou and Tessone}]{CHEN2022}
\bibinfo{author}{Chen, B.L.}, \bibinfo{author}{Jiang, W.X.}, \bibinfo{author}{Yu, Y.T.}, \bibinfo{author}{Zhou, L.}, \bibinfo{author}{Tessone, C.J.}, \bibinfo{year}{2022}.
\newblock \bibinfo{title}{Graph embedding based ant colony optimization for negative influence propagation suppression under cost constraints}.
\newblock \bibinfo{journal}{Swarm and Evolutionary Computation} \bibinfo{volume}{72}, \bibinfo{pages}{101102}.
\newblock \URLprefix \url{https://www.sciencedirect.com/science/article/pii/S2210650222000724}, \DOIprefix\doi{https://doi.org/10.1016/j.swevo.2022.101102}.
\bibitem[{Chen et~al.(2020)Chen, Wang, Wang and Kuo}]{Fenxiao2020}
\bibinfo{author}{Chen, F.}, \bibinfo{author}{Wang, Y.C.}, \bibinfo{author}{Wang, B.}, \bibinfo{author}{Kuo, C.C.}, \bibinfo{year}{2020}.
\newblock \bibinfo{title}{Graph representation learning: a survey}.
\newblock \bibinfo{journal}{APSIPA Transactions on Signal and Information Processing} \bibinfo{volume}{9}.
\newblock \DOIprefix\doi{10.1017/ATSIP.2020.13}.
\bibitem[{Dai et~al.(2017)Dai, Khalil, Zhang, Dilkina and Song}]{dai2018}
\bibinfo{author}{Dai, H.}, \bibinfo{author}{Khalil, E.B.}, \bibinfo{author}{Zhang, Y.}, \bibinfo{author}{Dilkina, B.}, \bibinfo{author}{Song, L.}, \bibinfo{year}{2017}.
\newblock \bibinfo{title}{Learning combinatorial optimization algorithms over graphs}, in: \bibinfo{booktitle}{Proceedings of the 31st International Conference on Neural Information Processing Systems}, \bibinfo{publisher}{Curran Associates Inc.}, \bibinfo{address}{Red Hook, NY, USA}. p. \bibinfo{pages}{6351–6361}.
\bibitem[{Dolan and Moré(2002)}]{Dolan2002}
\bibinfo{author}{Dolan, E.D.}, \bibinfo{author}{Moré, J.J.}, \bibinfo{year}{2002}.
\newblock \bibinfo{title}{Benchmarking optimization software with performance profiles}.
\newblock \bibinfo{journal}{Mathematical Programming} \bibinfo{volume}{91}, \bibinfo{pages}{201--213}.
\newblock \URLprefix \url{https://doi.org/10.1007/s101070100263}, \DOIprefix\doi{10.1007/s101070100263}.
\bibitem[{Eades(1991)}]{eades1991preserving}
\bibinfo{author}{Eades, P.}, \bibinfo{year}{1991}.
\newblock \bibinfo{title}{Preserving the Mental Map of a Diagram}.
\newblock IIAS-RR-, \bibinfo{publisher}{International Institute for Advanced Study of Social Information Science, Fujitsu Limited}.
\newblock \URLprefix \url{https://books.google.com.br/books?id=6v1BcgAACAAJ}.
\bibitem[{Feo and Resende(1995)}]{grasp}
\bibinfo{author}{Feo, T.}, \bibinfo{author}{Resende, M.}, \bibinfo{year}{1995}.
\newblock \bibinfo{title}{Greedy randomized adaptive search procedures}.
\newblock \bibinfo{journal}{Journal of Global Optimization} \bibinfo{volume}{6}, \bibinfo{pages}{109--133}.
\newblock \DOIprefix\doi{10.1007/BF01096763}.
\bibitem[{Gasse et~al.(2019)Gasse, Ch\'{e}telat, Ferroni, Charlin and Lodi}]{gasse2019}
\bibinfo{author}{Gasse, M.}, \bibinfo{author}{Ch\'{e}telat, D.}, \bibinfo{author}{Ferroni, N.}, \bibinfo{author}{Charlin, L.}, \bibinfo{author}{Lodi, A.}, \bibinfo{year}{2019}.
\newblock \bibinfo{title}{Exact combinatorial optimization with graph convolutional neural networks}. \bibinfo{publisher}{Curran Associates Inc.}, \bibinfo{address}{Red Hook, NY, USA}.
\bibitem[{Giovannangeli et~al.(2021)Giovannangeli, Lalanne, Auber, Giot and Bourqui}]{Giovannangeli2024}
\bibinfo{author}{Giovannangeli, L.}, \bibinfo{author}{Lalanne, F.}, \bibinfo{author}{Auber, D.}, \bibinfo{author}{Giot, R.}, \bibinfo{author}{Bourqui, R.}, \bibinfo{year}{2021}.
\newblock \bibinfo{title}{Deep neural network for drawing networks, {(DNN)\textsuperscript{2}}}, in: \bibinfo{editor}{Purchase, H.C.}, \bibinfo{editor}{Rutter, I.} (Eds.), \bibinfo{booktitle}{Graph Drawing and Network Visualization}, \bibinfo{publisher}{Springer}, \bibinfo{address}{Cham}. pp. \bibinfo{pages}{375--390}.
\bibitem[{Glover(1986)}]{Glover_Tabu}
\bibinfo{author}{Glover, F.}, \bibinfo{year}{1986}.
\newblock \bibinfo{title}{Future paths for integer programming and links to artificial intelligence}.
\newblock \bibinfo{journal}{Computers \& Operations Research} \bibinfo{volume}{13}, \bibinfo{pages}{533--549}.
\newblock \URLprefix \url{https://www.sciencedirect.com/science/article/pii/0305054886900481}, \DOIprefix\doi{https://doi.org/10.1016/0305-0548(86)90048-1}. \bibinfo{note}{applications of Integer Programming}.
\bibitem[{Goyal and Ferrara(2018)}]{Goyal:04}
\bibinfo{author}{Goyal, P.}, \bibinfo{author}{Ferrara, E.}, \bibinfo{year}{2018}.
\newblock \bibinfo{title}{Graph embedding techniques, applications, and performance: A survey}.
\newblock \bibinfo{journal}{Knowledge-Based Systems} \bibinfo{volume}{151}, \bibinfo{pages}{78–94}.
\newblock \URLprefix \url{http://dx.doi.org/10.1016/j.knosys.2018.03.022}, \DOIprefix\doi{10.1016/j.knosys.2018.03.022}.
\bibitem[{Grover and Leskovec(2016)}]{Grover}
\bibinfo{author}{Grover, A.}, \bibinfo{author}{Leskovec, J.}, \bibinfo{year}{2016}.
\newblock \bibinfo{title}{node2vec: Scalable feature learning for networks}, in: \bibinfo{booktitle}{Proceedings of the 22nd ACM SIGKDD International Conference on Knowledge Discovery and Data Mining}, \bibinfo{publisher}{Association for Computing Machinery}, \bibinfo{address}{New York, NY, USA}. p. \bibinfo{pages}{855–864}.
\newblock \URLprefix \url{https://doi.org/10.1145/2939672.2939754}, \DOIprefix\doi{10.1145/2939672.2939754}.
\bibitem[{G{\"u}nther et~al.(2001)G{\"u}nther, Sch{\"o}nfeld, Becker and Molitor}]{Gunther}
\bibinfo{author}{G{\"u}nther, W.}, \bibinfo{author}{Sch{\"o}nfeld, R.}, \bibinfo{author}{Becker, B.}, \bibinfo{author}{Molitor, P.}, \bibinfo{year}{2001}.
\newblock \bibinfo{title}{k-layer straightline crossing minimization by speeding up sifting}, in: \bibinfo{editor}{Marks, J.} (Ed.), \bibinfo{booktitle}{Graph Drawing}, \bibinfo{publisher}{Springer Berlin Heidelberg}, \bibinfo{address}{Berlin, Heidelberg}. pp. \bibinfo{pages}{253--258}.
\bibitem[{{Gurobi Optimization, LLC}(2023)}]{gurobi}
\bibinfo{author}{{Gurobi Optimization, LLC}}, \bibinfo{year}{2023}.
\newblock \bibinfo{title}{{Gurobi Optimizer Reference Manual}}.
\newblock \URLprefix \url{https://www.gurobi.com}.
\bibitem[{Hamilton(2020)}]{Hamiltonbook}
\bibinfo{author}{Hamilton, W.L.}, \bibinfo{year}{2020}.
\newblock \bibinfo{title}{Graph representation learning}.
\newblock \bibinfo{journal}{Synthesis Lectures on Artificial Intelligence and Machine Learning} \bibinfo{volume}{14}, \bibinfo{pages}{1--159}.
\bibitem[{Hamilton et~al.(2017)Hamilton, Ying and Leskovec}]{hamilton2018representation2.4}
\bibinfo{author}{Hamilton, W.L.}, \bibinfo{author}{Ying, R.}, \bibinfo{author}{Leskovec, J.}, \bibinfo{year}{2017}.
\newblock \bibinfo{title}{Representation learning on graphs: Methods and applications}.
\newblock \bibinfo{journal}{IEEE Data Eng. Bull.} \bibinfo{volume}{40}, \bibinfo{pages}{52--74}.
\newblock \URLprefix \url{https://api.semanticscholar.org/CorpusID:3215337}.
\bibitem[{Hoang et~al.(2023)Hoang, Jeon, You, Yoon, Jung and Lee}]{Hoang}
\bibinfo{author}{Hoang, V.T.}, \bibinfo{author}{Jeon, H.J.}, \bibinfo{author}{You, E.S.}, \bibinfo{author}{Yoon, Y.}, \bibinfo{author}{Jung, S.}, \bibinfo{author}{Lee, O.J.}, \bibinfo{year}{2023}.
\newblock \bibinfo{title}{Graph representation learning and its applications: A survey}.
\newblock \bibinfo{journal}{Sensors} \bibinfo{volume}{23}.
\newblock \URLprefix \url{https://www.mdpi.com/1424-8220/23/8/4168}, \DOIprefix\doi{10.3390/s23084168}.
\bibitem[{Ju et~al.(2024)Ju, Fang, Gu, Liu, Long, Qiao, Qin, Shen, Sun, Xiao, Yang, Yuan, Zhao, Wang, Luo and Zhang}]{Wei_Ju_2024}
\bibinfo{author}{Ju, W.}, \bibinfo{author}{Fang, Z.}, \bibinfo{author}{Gu, Y.}, \bibinfo{author}{Liu, Z.}, \bibinfo{author}{Long, Q.}, \bibinfo{author}{Qiao, Z.}, \bibinfo{author}{Qin, Y.}, \bibinfo{author}{Shen, J.}, \bibinfo{author}{Sun, F.}, \bibinfo{author}{Xiao, Z.}, \bibinfo{author}{Yang, J.}, \bibinfo{author}{Yuan, J.}, \bibinfo{author}{Zhao, Y.}, \bibinfo{author}{Wang, Y.}, \bibinfo{author}{Luo, X.}, \bibinfo{author}{Zhang, M.}, \bibinfo{year}{2024}.
\newblock \bibinfo{title}{A comprehensive survey on deep graph representation learning}.
\newblock \bibinfo{journal}{Neural Networks} \bibinfo{volume}{173}, \bibinfo{pages}{106207}.
\newblock \URLprefix \url{https://www.sciencedirect.com/science/article/pii/S089360802400131X}, \DOIprefix\doi{https://doi.org/10.1016/j.neunet.2024.106207}.
\bibitem[{J\"{u}nger et~al.(1997)J\"{u}nger, Lee, Mutzel and Odenthal}]{Junger1997}
\bibinfo{author}{J\"{u}nger, M.}, \bibinfo{author}{Lee, E.K.}, \bibinfo{author}{Mutzel, P.}, \bibinfo{author}{Odenthal, T.}, \bibinfo{year}{1997}.
\newblock \bibinfo{title}{A polyhedral approach to the multi-layer crossing minimization problem}, in: \bibinfo{booktitle}{Proceedings of the 5th International Symposium on Graph Drawing}, \bibinfo{publisher}{Springer-Verlag}, \bibinfo{address}{Berlin, Heidelberg}. p. \bibinfo{pages}{13–24}.
\bibitem[{Kallestad et~al.(2023)Kallestad, Hasibi, Hemmati and S{\"o}rensen}]{KALLESTAD2023446}
\bibinfo{author}{Kallestad, J.}, \bibinfo{author}{Hasibi, R.}, \bibinfo{author}{Hemmati, A.}, \bibinfo{author}{S{\"o}rensen, K.}, \bibinfo{year}{2023}.
\newblock \bibinfo{title}{A general deep reinforcement learning hyperheuristic framework for solving combinatorial optimization problems}.
\newblock \bibinfo{journal}{European Journal of Operational Research} \bibinfo{volume}{309}, \bibinfo{pages}{446--468}.
\newblock \URLprefix \url{https://www.sciencedirect.com/science/article/pii/S037722172300036X}, \DOIprefix\doi{10.1016/j.ejor.2023.01.017}.
\bibitem[{Karimi~Mamaghan et~al.(2021)Karimi~Mamaghan, Mohammadi, Meyer, Karimi~Mamaghan and Talbi}]{Karimi}
\bibinfo{author}{Karimi~Mamaghan, M.}, \bibinfo{author}{Mohammadi, M.}, \bibinfo{author}{Meyer, P.}, \bibinfo{author}{Karimi~Mamaghan, A.M.}, \bibinfo{author}{Talbi, E.G.}, \bibinfo{year}{2021}.
\newblock \bibinfo{title}{Machine learning at the service of meta-heuristics for solving combinatorial optimization problems: A state-of-the-art}.
\newblock \bibinfo{journal}{European Journal of Operational Research} \bibinfo{volume}{296}.
\newblock \DOIprefix\doi{10.1016/j.ejor.2021.04.032}.
\bibitem[{Kaufmann and Wagner(2001)}]{kaufmann3.0}
\bibinfo{author}{Kaufmann, M.}, \bibinfo{author}{Wagner, D.}, \bibinfo{year}{2001}.
\newblock \bibinfo{title}{Drawing Graphs: Methods and Models}. volume \bibinfo{volume}{2025}.
\newblock \bibinfo{publisher}{Springer}, \bibinfo{address}{Berlin}.
\bibitem[{Khoshraftar and An(2024)}]{shima2024}
\bibinfo{author}{Khoshraftar, S.}, \bibinfo{author}{An, A.}, \bibinfo{year}{2024}.
\newblock \bibinfo{title}{A survey on graph representation learning methods}.
\newblock \bibinfo{journal}{ACM Trans. Intell. Syst. Technol.} \bibinfo{volume}{15}.
\newblock \URLprefix \url{https://doi.org/10.1145/3633518}, \DOIprefix\doi{10.1145/3633518}.
\bibitem[{Laguna et~al.(2025)Laguna, Martí, Martínez-Gavara, Pérez-Peló and Resende}]{LAGUNA2025}
\bibinfo{author}{Laguna, M.}, \bibinfo{author}{Martí, R.}, \bibinfo{author}{Martínez-Gavara, A.}, \bibinfo{author}{Pérez-Peló, S.}, \bibinfo{author}{Resende, M.G.}, \bibinfo{year}{2025}.
\newblock \bibinfo{title}{Greedy randomized adaptive search procedures with path relinking. an analytical review of designs and implementations}.
\newblock \bibinfo{journal}{European Journal of Operational Research} \URLprefix \url{https://www.sciencedirect.com/science/article/pii/S0377221725001456}, \DOIprefix\doi{10.1016/j.ejor.2025.02.022}.
\bibitem[{Li et~al.(2018)Li, Chen and Koltun}]{li2018}
\bibinfo{author}{Li, Z.}, \bibinfo{author}{Chen, Q.}, \bibinfo{author}{Koltun, V.}, \bibinfo{year}{2018}.
\newblock \bibinfo{title}{Combinatorial optimization with graph convolutional networks and guided tree search}, in: \bibinfo{booktitle}{Proceedings of the 32nd International Conference on Neural Information Processing Systems}, \bibinfo{publisher}{Curran Associates Inc.}, \bibinfo{address}{Red Hook, NY, USA}. p. \bibinfo{pages}{537–546}.
\bibitem[{Liu et~al.(2022)Liu, Chen and Weiszer}]{LIU2022}
\bibinfo{author}{Liu, S.}, \bibinfo{author}{Chen, J.}, \bibinfo{author}{Weiszer, M.}, \bibinfo{year}{2022}.
\newblock \bibinfo{title}{Multi-objective multigraph a* search with learning heuristics based on node metrics and graph embedding}, in: \bibinfo{booktitle}{2022 IEEE 11th International Conference on Intelligent Systems (IS)}, pp. \bibinfo{pages}{1--8}.
\newblock \DOIprefix\doi{10.1109/IS57118.2022.10019653}.
\bibitem[{M{\"a}kinen and Siirtola(2005)}]{Makinen}
\bibinfo{author}{M{\"a}kinen, E.}, \bibinfo{author}{Siirtola, H.}, \bibinfo{year}{2005}.
\newblock \bibinfo{title}{The barycenter heuristic and the reorderable matrix.}
\newblock \bibinfo{journal}{Informatica (Slovenia)} \bibinfo{volume}{29}, \bibinfo{pages}{357--364}.
\bibitem[{Martí et~al.(2018)Martí, Martínez-Gavara, Sánchez-Oro and Duarte}]{MARTI20181}
\bibinfo{author}{Martí, R.}, \bibinfo{author}{Martínez-Gavara, A.}, \bibinfo{author}{Sánchez-Oro, J.}, \bibinfo{author}{Duarte, A.}, \bibinfo{year}{2018}.
\newblock \bibinfo{title}{Tabu search for the dynamic bipartite drawing problem}.
\newblock \bibinfo{journal}{Computers \& Operations Research} \bibinfo{volume}{91}, \bibinfo{pages}{1--12}.
\newblock \URLprefix \url{https://www.sciencedirect.com/science/article/pii/S0305054817302745}, \DOIprefix\doi{https://doi.org/10.1016/j.cor.2017.10.011}.
\bibitem[{Mart\'i
 and Estruch(2001)}]{MARTI20011287}
\bibinfo{author}{Mart\'i
, R.}, \bibinfo{author}{Estruch, V.}, \bibinfo{year}{2001}.
\newblock \bibinfo{title}{Incremental bipartite drawing problem}.
\newblock \bibinfo{journal}{Computers \& Operations Research} \bibinfo{volume}{28}, \bibinfo{pages}{1287--1298}.
\newblock \URLprefix \url{https://www.sciencedirect.com/science/article/pii/S030505480000040X}, \DOIprefix\doi{https://doi.org/10.1016/S0305-0548(00)00040-X}.
\bibitem[{Maćkiewicz and Ratajczak(1993)}]{pca}
\bibinfo{author}{Maćkiewicz, A.}, \bibinfo{author}{Ratajczak, W.}, \bibinfo{year}{1993}.
\newblock \bibinfo{title}{Principal components analysis (pca)}.
\newblock \bibinfo{journal}{Computers \& Geosciences} \bibinfo{volume}{19}, \bibinfo{pages}{303--342}.
\newblock \URLprefix \url{https://www.sciencedirect.com/science/article/pii/009830049390090R}, \DOIprefix\doi{https://doi.org/10.1016/0098-3004(93)90090-R}.
\bibitem[{Napoletano et~al.(2019)Napoletano, Mart\'inez-Gavara, Festa, Pastore and Mart\'i}]{napoletano2019}
\bibinfo{author}{Napoletano, A.}, \bibinfo{author}{Mart\'inez-Gavara, A.}, \bibinfo{author}{Festa, P.}, \bibinfo{author}{Pastore, T.}, \bibinfo{author}{Mart\'i, R.}, \bibinfo{year}{2019}.
\newblock \bibinfo{title}{Heuristics for the constrained incremental graph drawing problem}.
\newblock \bibinfo{journal}{European Journal of Operational Research} \bibinfo{volume}{274}, \bibinfo{pages}{710--729}.
\bibitem[{Nascimento et~al.(2010)Nascimento, Resende and Toledo}]{NASCIMENTO2010747}
\bibinfo{author}{Nascimento, M.C.}, \bibinfo{author}{Resende, M.G.}, \bibinfo{author}{Toledo, F.M.}, \bibinfo{year}{2010}.
\newblock \bibinfo{title}{Grasp heuristic with path-relinking for the multi-plant capacitated lot sizing problem}.
\newblock \bibinfo{journal}{European Journal of Operational Research} \bibinfo{volume}{200}, \bibinfo{pages}{747--754}.
\newblock \URLprefix \url{https://www.sciencedirect.com/science/article/pii/S0377221709000459}, \DOIprefix\doi{10.1016/j.ejor.2009.01.047}.
\bibitem[{Ou et~al.(2016)Ou, Cui, Pei, Zhang and Zhu}]{Mingdong}
\bibinfo{author}{Ou, M.}, \bibinfo{author}{Cui, P.}, \bibinfo{author}{Pei, J.}, \bibinfo{author}{Zhang, Z.}, \bibinfo{author}{Zhu, W.}, \bibinfo{year}{2016}.
\newblock \bibinfo{title}{Asymmetric transitivity preserving graph embedding}, in: \bibinfo{booktitle}{Proceedings of the 22nd ACM SIGKDD International Conference on Knowledge Discovery and Data Mining}, \bibinfo{publisher}{Association for Computing Machinery}, \bibinfo{address}{New York, NY, USA}. p. \bibinfo{pages}{1105–1114}.
\newblock \URLprefix \url{https://doi.org/10.1145/2939672.2939751}, \DOIprefix\doi{10.1145/2939672.2939751}.
\bibitem[{Peng et~al.(2020)Peng, Liu, Lü, Martí and Ding}]{PENG2020183}
\bibinfo{author}{Peng, B.}, \bibinfo{author}{Liu, D.}, \bibinfo{author}{Lü, Z.}, \bibinfo{author}{Martí, R.}, \bibinfo{author}{Ding, J.}, \bibinfo{year}{2020}.
\newblock \bibinfo{title}{Adaptive memory programming for the dynamic bipartite drawing problem}.
\newblock \bibinfo{journal}{Information Sciences} \bibinfo{volume}{517}, \bibinfo{pages}{183--197}.
\newblock \URLprefix \url{https://www.sciencedirect.com/science/article/pii/S0020025519312046}, \DOIprefix\doi{https://doi.org/10.1016/j.ins.2019.12.077}.
\bibitem[{Peng et~al.(2024)Peng, Wang, Liu, Su, Lü and Glover}]{PENG2024121477}
\bibinfo{author}{Peng, B.}, \bibinfo{author}{Wang, S.}, \bibinfo{author}{Liu, D.}, \bibinfo{author}{Su, Z.}, \bibinfo{author}{Lü, Z.}, \bibinfo{author}{Glover, F.}, \bibinfo{year}{2024}.
\newblock \bibinfo{title}{Solving the incremental graph drawing problem by multiple neighborhood solution-based tabu search algorithm}.
\newblock \bibinfo{journal}{Expert Systems with Applications} \bibinfo{volume}{237}, \bibinfo{pages}{121477}.
\newblock \URLprefix \url{https://www.sciencedirect.com/science/article/pii/S0957417423019796}, \DOIprefix\doi{https://doi.org/10.1016/j.eswa.2023.121477}.
\bibitem[{Peres and Castelli(2021)}]{Peres}
\bibinfo{author}{Peres, F.}, \bibinfo{author}{Castelli, M.}, \bibinfo{year}{2021}.
\newblock \bibinfo{title}{Combinatorial optimization problems and metaheuristics: Review, challenges, design, and development}.
\newblock \bibinfo{journal}{Applied Sciences} \bibinfo{volume}{11}.
\newblock \DOIprefix\doi{10.3390/app11146449}.
\bibitem[{Perozzi et~al.(2014)Perozzi, Al-Rfou and Skiena}]{Perozzi2014}
\bibinfo{author}{Perozzi, B.}, \bibinfo{author}{Al-Rfou, R.}, \bibinfo{author}{Skiena, S.}, \bibinfo{year}{2014}.
\newblock \bibinfo{title}{Deepwalk: Online learning of social representations}, in: \bibinfo{booktitle}{Proceedings of the 20th ACM SIGKDD International Conference on Knowledge Discovery and Data Mining}, \bibinfo{publisher}{Association for Computing Machinery}, \bibinfo{address}{New York, NY, USA}. p. \bibinfo{pages}{701–710}.
\newblock \URLprefix \url{https://doi.org/10.1145/2623330.2623732}, \DOIprefix\doi{10.1145/2623330.2623732}.
\bibitem[{Resende and Ribeiro(2005)}]{ResendeRibeiro}
\bibinfo{author}{Resende, M.}, \bibinfo{author}{Ribeiro, C.}, \bibinfo{year}{2005}.
\newblock \bibinfo{title}{Grasp with path-relinking: Recent advances and applications}.
\newblock \bibinfo{journal}{Operations Research/ Computer Science Interfaces Series} \bibinfo{volume}{32}.
\newblock \DOIprefix\doi{10.1007/0-387-25383-1_2}.
\bibitem[{S\'{a}nchez-Oro et~al.(2017)S\'{a}nchez-Oro, Mart\'{\i}nez-Gavara, Laguna, Mart\'{\i} and Duarte}]{Sanchez2017}
\bibinfo{author}{S\'{a}nchez-Oro, J.}, \bibinfo{author}{Mart\'{\i}nez-Gavara, A.}, \bibinfo{author}{Laguna, M.}, \bibinfo{author}{Mart\'{\i}, R.}, \bibinfo{author}{Duarte, A.}, \bibinfo{year}{2017}.
\newblock \bibinfo{title}{Variable neighborhood scatter search for the incremental graph drawing problem}.
\newblock \bibinfo{journal}{Comput. Optim. Appl.} \bibinfo{volume}{68}, \bibinfo{pages}{775–797}.
\newblock \URLprefix \url{https://doi.org/10.1007/s10589-017-9926-5}, \DOIprefix\doi{10.1007/s10589-017-9926-5}.
\bibitem[{Song et~al.(2019)Song, Triguero and {\"O}zcan}]{Song2019_1.3}
\bibinfo{author}{Song, H.}, \bibinfo{author}{Triguero, I.}, \bibinfo{author}{{\"O}zcan, E.}, \bibinfo{year}{2019}.
\newblock \bibinfo{title}{A review on the self and dual interactions between machine learning and optimization}.
\newblock \bibinfo{journal}{Progress in Artificial Intelligence} \bibinfo{volume}{8}, \bibinfo{pages}{143--165}.
\bibitem[{{Souza Almeida} et~al.(2022){Souza Almeida}, Goerlandt, Pelot and S{\"o}rensen}]{SOUZAALMEIDA2022105804}
\bibinfo{author}{{Souza Almeida}, L.}, \bibinfo{author}{Goerlandt, F.}, \bibinfo{author}{Pelot, R.}, \bibinfo{author}{S{\"o}rensen, K.}, \bibinfo{year}{2022}.
\newblock \bibinfo{title}{A greedy randomized adaptive search procedure (grasp) for the multi-vehicle prize collecting arc routing for connectivity problem}.
\newblock \bibinfo{journal}{Computers \& Operations Research} \bibinfo{volume}{143}, \bibinfo{pages}{105804}.
\newblock \URLprefix \url{https://www.sciencedirect.com/science/article/pii/S0305054822000910}, \DOIprefix\doi{10.1016/j.cor.2022.105804}.
\bibitem[{Sugiyama et~al.(1981)Sugiyama, Tagawa and Toda}]{Sugiyama3.9}
\bibinfo{author}{Sugiyama, K.}, \bibinfo{author}{Tagawa, S.}, \bibinfo{author}{Toda, M.}, \bibinfo{year}{1981}.
\newblock \bibinfo{title}{Methods for visual understanding of hierarchical system structures}.
\newblock \bibinfo{journal}{IEEE Transactions on Systems, Man, and Cybernetics} \bibinfo{volume}{11}, \bibinfo{pages}{109--125}.
\bibitem[{Talbi(2016)}]{RePEc_Talbi}
\bibinfo{author}{Talbi, E.}, \bibinfo{year}{2016}.
\newblock \bibinfo{title}{{Combining metaheuristics with mathematical programming, constraint programming and machine learning}}.
\newblock \bibinfo{journal}{Annals of Operations Research} \bibinfo{volume}{240}, \bibinfo{pages}{171--215}.
\newblock \DOIprefix\doi{10.1007/s10479-015-2034-y}. \bibinfo{note}{available at : \url{https://ideas.repec.org/a/spr/annopr/v240y2016i1d10.1007_s10479-015-2034-y.html} Acessed on: May $05^{th}$, 2024.}
\bibitem[{Talbi(2021)}]{Talbi2021}
\bibinfo{author}{Talbi, E.G.}, \bibinfo{year}{2021}.
\newblock \bibinfo{title}{Machine learning into metaheuristics: A survey and taxonomy}.
\newblock \bibinfo{journal}{ACM Computing Surveys} \bibinfo{volume}{54}, \bibinfo{pages}{1--32}.
\newblock \DOIprefix\doi{10.1145/3459664}.
\bibitem[{Tamassia and Tollis(1995)}]{tamassia1995graph}
\bibinfo{author}{Tamassia, R.}, \bibinfo{author}{Tollis, I.G.}, \bibinfo{year}{1995}.
\newblock \bibinfo{title}{Graph Drawing: DIMACS International Workshop, GD '94, Princeton, New Jersey, USA, October 10-12, 1994 : Proceedings}.
\newblock Lecture notes in computer science, \bibinfo{publisher}{Springer-Verlag}.
\newblock \URLprefix \url{https://books.google.com.br/books?id=kaFTvwEACAAJ}.
\bibitem[{Tang and Liu(2011)}]{Tang}
\bibinfo{author}{Tang, L.}, \bibinfo{author}{Liu, H.}, \bibinfo{year}{2011}.
\newblock \bibinfo{title}{Leveraging social media networks for classification}.
\newblock \bibinfo{journal}{Data Min. Knowl. Discov.} \bibinfo{volume}{23}, \bibinfo{pages}{447--478}.
\newblock \DOIprefix\doi{10.1007/s10618-010-0210-x}.
\bibitem[{Tiezz et~al.(2024)Tiezz, Ciravegna and Gori}]{Tiezzi2022}
\bibinfo{author}{Tiezz, M.}, \bibinfo{author}{Ciravegna, G.}, \bibinfo{author}{Gori, M.}, \bibinfo{year}{2024}.
\newblock \bibinfo{title}{Graph neural networks for graph drawing}.
\newblock \bibinfo{journal}{IEEE Transactions on Neural Networks and Learning Systems} \bibinfo{volume}{35}, \bibinfo{pages}{4668--4681}.
\bibitem[{Wang et~al.(2021)Wang, Yen, Hu and Shen}]{Wang2021_}
\bibinfo{author}{Wang, X.}, \bibinfo{author}{Yen, K.}, \bibinfo{author}{Hu, Y.}, \bibinfo{author}{Shen, H.W.}, \bibinfo{year}{2021}.
\newblock \bibinfo{title}{{ DeepGD: A Deep Learning Framework for Graph Drawing Using GNN }}.
\newblock \bibinfo{journal}{IEEE Computer Graphics and Applications} \bibinfo{volume}{41}, \bibinfo{pages}{32--44}.
\newblock \DOIprefix\doi{10.1109/MCG.2021.3093908}. \bibinfo{note}{dOI: \url{https://doi.ieeecomputersociety.org/10.1109/MCG.2021.3093908}}.
\bibitem[{Wang et~al.(2020)Wang, Jin, Wang, Cui, Ma and Qu}]{wang3.43}
\bibinfo{author}{Wang, Y.}, \bibinfo{author}{Jin, Z.}, \bibinfo{author}{Wang, Q.}, \bibinfo{author}{Cui, W.}, \bibinfo{author}{Ma, T.}, \bibinfo{author}{Qu, H.}, \bibinfo{year}{2020}.
\newblock \bibinfo{title}{Deepdrawing: a deep learning approach to graph drawing}.
\newblock \bibinfo{journal}{IEEE Transactions on Visualization and Computer Graphics} \bibinfo{volume}{26}, \bibinfo{pages}{676--686}.
\bibitem[{Ware et~al.(2002)Ware, Purchase, Colpoys and McGill}]{ware3.4}
\bibinfo{author}{Ware, C.}, \bibinfo{author}{Purchase, H.}, \bibinfo{author}{Colpoys, L.}, \bibinfo{author}{McGill, M.}, \bibinfo{year}{2002}.
\newblock \bibinfo{title}{Cognitive measurements of graph aesthetics}.
\newblock \bibinfo{journal}{Information Visualization} \bibinfo{volume}{1}, \bibinfo{pages}{103--110}.
\newblock \URLprefix \url{https://doi.org/10.1057/palgrave.ivs.9500013}, \DOIprefix\doi{10.1057/palgrave.ivs.9500013}, \href{http://arxiv.org/abs/https://doi.org/10.1057/palgrave.ivs.9500013}{{\tt arXiv:https://doi.org/10.1057/palgrave.ivs.9500013}}.
\bibitem[{Ye et~al.(2023)Ye, Wang, Cao, Liang and Li}]{ye2023}
\bibinfo{author}{Ye, H.}, \bibinfo{author}{Wang, J.}, \bibinfo{author}{Cao, Z.}, \bibinfo{author}{Liang, H.}, \bibinfo{author}{Li, Y.}, \bibinfo{year}{2023}.
\newblock \bibinfo{title}{Deepaco: neural-enhanced ant systems for combinatorial optimization}, in: \bibinfo{booktitle}{Proceedings of the 37th International Conference on Neural Information Processing Systems}, \bibinfo{publisher}{Curran Associates Inc.}, \bibinfo{address}{Red Hook, NY, USA}.

\end{thebibliography}
\end{document}